\pdfoutput=1

\documentclass[11pt]{article}
\usepackage[table]{xcolor}

\usepackage[final]{acl}

\usepackage{times}
\usepackage{latexsym}

\usepackage[T1]{fontenc}

\usepackage[utf8]{inputenc}

\usepackage{microtype}

\usepackage{inconsolata}

\usepackage{graphicx}

\usepackage{amsmath}
\usepackage{bm}
\usepackage{booktabs}
\usepackage{multirow}
\usepackage{physics}
\usepackage{makecell}
\usepackage{mathtools}
\usepackage{amsmath}
\usepackage{cleveref}
\Crefname{section}{\S}{\S\S}
\usepackage{amssymb}
\usepackage{hyperref}
\usepackage{fontawesome}

\usepackage[framemethod=TikZ]{mdframed}
\newcommand{\graybox}[1]{\begin{mdframed}[
    backgroundcolor=black!5,
    topline=false, bottomline=false, rightline=false, leftline=false,
    innertopmargin=0.5em,
    innerleftmargin=0.5em,
    innerbottommargin=0.5em,
    innerrightmargin=0.5em,
]{#1}
\end{mdframed}
}

\title{Can Input Attributions Explain Inductive Reasoning \\ in In-Context Learning?}

\author{Mengyu Ye${}^{1}$ \hspace{1em} 
        Tatsuki Kuribayashi${}^{2,1}$ \hspace{1em}
        Goro Kobayashi\thanks{Currently affiliated with Preferred Networks, Inc.}${}^{1,3}$ \hspace{1em} 
        Jun Suzuki${}^{1,3}$ \hspace{1em} \\
        ${}^{1}$Tohoku University \hspace{1em}
        ${}^{2}$MBZUAI \hspace{1em}
        ${}^{3}$RIKEN \\
        \texttt{\{ye.mengyu.s1, goro.koba\}@dc.tohoku.ac.jp} \\
        \texttt{tatsuki.kuribayashi@mbzuai.ac.ae} \hspace{1em}
        \texttt{jun.suzuki@tohoku.ac.jp}
        }

\begin{document}
\maketitle
\begin{abstract}
Interpreting the internal process of neural models has long been a challenge.
This challenge remains relevant in the era of large language models (LLMs) and in-context learning (ICL); for example, ICL poses a new issue of interpreting which example in the few-shot examples contributed to identifying/solving the task.
To this end, in this paper, we design synthetic diagnostic tasks of inductive reasoning, inspired by the generalization tests typically adopted in psycholinguistics.
Here, most in-context examples are ambiguous w.r.t. their underlying rule, and one critical example disambiguates it. 
The question is whether conventional input attribution (IA) methods can track such a reasoning process, i.e., identify the influential example, in ICL. 
Our experiments provide several practical findings; for example, a certain simple IA method works the best, and the larger the model, the generally harder it is to interpret the ICL with gradient-based IA methods.

\large{\faicon{github}}
\hspace{.25em}
\parbox{\dimexpr\linewidth-6\fboxsep-2\fboxrule}{\sloppy \small \url{https://github.com/muyo8692/input-attribution-icl}}
\end{abstract}
\begin{figure}[t]
\centering
\includegraphics[width=\linewidth]{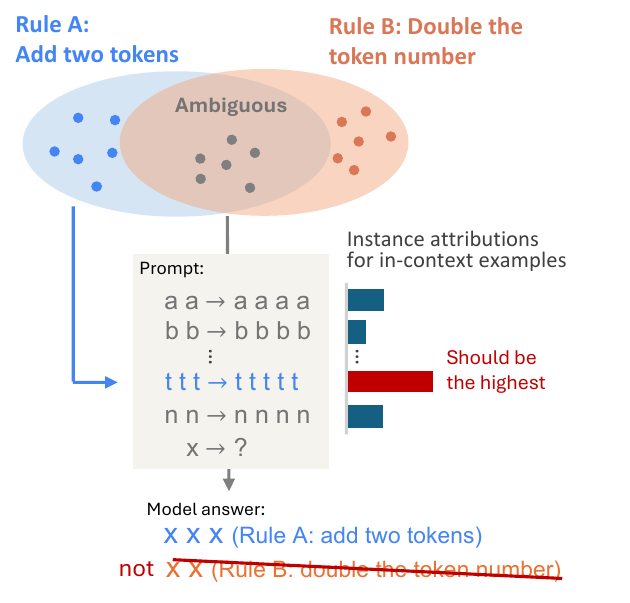}
\caption{Overview of our experimental setup. The majority of in-context examples (gray) are ambiguous, supporting either Rule A of adding two tokens or Rule B of doubling tokens. A single disambiguating example (blue) reveals that Rule A is correct. We investigate whether input attribution (IA) methods can track such an inductive reasoning process.}
\label{fig:fig1}
\end{figure}
\section{Introduction}
In the natural language processing (NLP) field, input attribution (IA) methods, e.g., relying on gradient norm~\cite{simonyan-etal-2014-saliency, li-etal-2016-visualizing}, have typically been employed to interpret input--output associations exploited by neural NLP models~\cite{vinyals-and-le-2015-neural, li-etal-2016-diversity}.
Recently, as large language models (LLMs) and mechanistic interpretability (MI) research~\cite{olah2020zoom,bereska2024mechanistic} has gained attention,  the research focus is leaning toward understanding the \textit{circuits} within LLMs by intervening in their internal representations and information flow.
Despite the enriched scope of research, such rapid progress has missed some intriguing questions bridging the IA and MI eras: in particular, \textit{do conventional IA methods still empirically work in the modern NLP setting, specifically with LLMs and in-context learning (ICL)?}

In this paper, we revisit IA methods in interpreting LLM-based ICL~\cite{brown-etal-2020-learners}.
Specifically, we assess how well IA methods can track the most influential example in a few-shot examples. %
This question is worth investigating for several reasons. 
First, input attribution would still serve as a necessary and sufficient explanation in typical practical cases; some users might simply seek which part of the context is heavily referred to by an LLM system rather than LLMs' internal processes identified by MI methods. 
Second, the modern NLP setting, specifically ICL, differs from the conventional settings where IA methods have been tested --- identifying the input-output association within a specific instance ($X_k$, $y_k$). 
In contrast, applying IA methods to the entire ICL input (few-shot examples) entails tracking the model's \textit{learning} process over these instances as well as the pure input-output association within a specific target instance.
This involves the interpretation of which \textit{example} among the demonstrations [$(X_1,y_1), \cdots, (X_{k-1},y_{k-1})$] contributed to identifying the targeted task/rule and then answering a target question $X_k$.
This is rather a type of instance-based interpretation of neural models~\cite{wachter-etal-2017-conterfactual, charpiat-etal-2019-input, hanawa-etal-2021-evaluation}, and it has been little explored such interpretation is feasible with IA methods.

To test IA methods in ICL, we introduce a test suite comprising controlled synthetic inductive reasoning tasks.
Otherwise, formally defining such informativeness and assessing IA methods is challenging, especially in a wild, natural setting; critical examples may not be unique~\cite{min-etal-2022-rethinking}, a gap might exist between faithfulness and plausibility perspectives~\cite{bastings-etal-2022-will}, and a model can rather rely on prior knowledge without using any input examples~\cite{liu-etal-2022-makes}.
Our task design, inspired by the poverty of the stimulus scenario~\cite{Wilson2006-qo,Perfors2011-mu,mccoy2018revisiting,mccoy-etal-2020-syntax,Yedetore2023-sw} or mixed signals generalization test~\cite{Warstadt2020-tk,Mueller2024-ae}, introduces one inherently unique \textit{aha} example in input demonstrations.
This \textit{aha} example, when paired with any of the other examples, triggers the identification of the underlying reasoning rule.
More specifically, most in-context examples are \textit{ambiguous} in the sense that they are compatible with several rules (e.g., adding two tokens or doubling the number of tokens, in Figure~\ref{fig:fig1}), and only one \textit{disambiguating} (\textit{aha}) example resolves the ambiguity and limits the correct rule to be unique (\texttt{ttt$\rightarrow$ttttt} disambiguates the rule to be \textit{adding} one in Figure~\ref{fig:fig1}). 
The question is whether such an informative example can be empirically caught by IA methods.

In our experiments, IA methods are compared to other common interpretability approaches relying on, such as attention weights and post-hoc explanation generations. 
The experimental results demonstrate:
\begin{itemize}
\setlength{\parskip}{0cm}
\setlength{\itemsep}{0.05cm}
    \item Gradient norm, the simplest IA method, frequently outperforms other interpretability methods (e.g., integrated gradient, attention weights), suggesting that the advantage of more recently proposed IA methods does not always generalize in interpreting ICL$\times$LLM.
    \item %
    We revealed that different interpretability methods exhibited different advantages in interpreting the ICL process, against scaling in several aspects.  For example, IA methods perform better in many-shot scenarios, whereas a post-hoc explanation generations work well on larger models.
    \item Our tested interpretability methods, including simple gradient norm, did not work stably across different tasks and models, posing their general limitations in interpreting ICL with IA methods. Some existing IA methods frequently failed even on a very simple associative recall task, and there is room to sophisticate previously developed interpretability tools to be suitable for LLMs.

\end{itemize}

\section{Preliminary}

\subsection{Input attribution (IA) methods}
\label{sec:method-def}
Input attribution (IA) methods are commonly-used techniques for interpreting and explaining the predictions of machine learning models~\cite[][etc.]{denil-etal-2014-extraction, li-etal-2016-visualizing, poerner-etal-2018-evaluating, arras-etal-2019-evaluating}. 
Specifically, IA methods determine how much each input feature contributes to a particular prediction; that is, given input tokens $X \coloneqq [x_1, \ldots, x_n]$ and output $y$, the IA methods yield the strength of contribution $S(x_i)$ of each input token $x_i \in X$ to the output $y$.
Note that the input $X$ in ICL consists of tokens of several in-context examples (\cref{sec:prompt-format}), and the answer to the target question is denoted as $y$.
We examine the following four representative IA methods in the ICL$\times$LLM context:

\paragraph{Input erasure (\textsc{IE})}
\textsc{IE} \cite{li-etal-2016-erasure} measures how impactful erasing a certain token $x_i$ from the input prompt is with respect to outputing $y_t$:
\begin{align}
    S_{\mathrm{IE}}(x_i, y_t; X) &= q(y_t \vert \bm X) - q(y_t \vert \bm X_{\neg i})
    \text{,}
    \label{eq:contrast-ie}
\end{align}
\noindent
where $\bm X \coloneqq [\bm x_1, \ldots, \bm x_n]$ denotes the sequence of input token embeddings, with each $\bm x_i \in \mathbb{R}^{d}$ being a $d$-dimensional vector corresponding to the $i$-th token in the input.
$\bm X_{\neg i} \coloneqq [\bm x_1, \ldots, \bm x_{i-1},\bm x_{i+1}, \ldots, \bm x_n]$ denotes the sequence of input token embeddings without $\bm x_i$.
We emulate this partial input $\bm X_{\neg i}$ by introducing an attention mask to zero-out the attention to $\bm x_{i}$ in every layer (thus, the original position information holds).
$q(y \vert \bm X)$ represents the model's prediction probability for the token $y_t$ given input $\bm X$.

\paragraph{Gradient norm (\textsc{GN})}
\textsc{GN}~\cite{simonyan-etal-2014-saliency, li-etal-2016-visualizing} calculates the attribution %
score for each input token $x_i$ by computing the L1 norm of its gradient of the target token $y_t$:
\begin{align}
    S_{\mathrm{GN}}(\bm x_i, y_t; \bm X) = {\lVert g(\bm x_i, y_t; \bm X) \rVert}_{\mathrm{L1}}
    \\
    g(\bm x_i, y_t; \bm X) = \nabla_{\bm x_i}q(y_t \vert \bm X)
    \text{,}
    \label{eq:contrast-gn}
\end{align}
\noindent
where $g(\bm x_i, y_t; \bm X) \in \mathbb{R}^d$ denotes the gradient of the prediction probability for $y_t$ with respect to $\bm x_i$, under the given input embedding sequence $\bm X$.

\paragraph{Input$\times$gradient (\textsc{I$\times$G})}
\textsc{I$\times$G} \cite{shrikumar-etal-2017-learning, denil-etal-2014-extraction} takes the dot product of a gradient with the respective token embedding $\bm x_i$:
\begin{equation}
    S_{\mathrm{I\times G}}(\bm x_i, y_t; \bm X) = g(\bm x_i, y_t; \bm X) \cdot \bm x_i
    \text{.}
\label{eq:contrast-ixg}
\end{equation}

\paragraph{Integrated gradients (\textsc{IG})}
\textsc{IG} \cite{sundararajan-etal-2017-axiomatic} is computed by accumulating gradients along a straight path from a baseline input $\bm X'$ to the actual input $\bm X$:
\begin{align}
    &S_{\mathrm{IG}}(\bm x_i, y_t; \bm X) = 
    \nonumber
    \\
    & \quad (\bm x_i - \bm x'_i) \times \int_{0}^1 \frac{\partial q(y_t \vert \bm X' + \alpha(\bm X - \bm X'))}{\partial \bm x_i} \, d\alpha
    \text{,}
    \label{eq:contrast-ig}
\end{align}
where $\bm X' \coloneqq [\bm x'_1, \ldots, \bm x'_n]$ denotes the sequence of baseline embeddings\footnote{
    We followed the common practice and employed a sequence of zero vectors as the baseline input. 
    We used an interpretability library \href{https://captum.ai/}{captum}~\cite{kokhlikyan2020captum} to calculate the IG score and keep all parameters as default.}, and $\alpha$ denotes the interpolation coefficient.
In practice, the integral is approximated using numerical integration with a finite number of steps.

\paragraph{Contrastive explanations}
For the \textsc{IE}, \textsc{GN}, and \textsc{I$\times$G} methods, we adopt a contrastive explanation setting, which \citet{yin-neubig-2022-interpreting} have shown to be quantitatively superior to the original non-contrastive setting.
IA methods in this setting measure how much an input token $x_i$ influences the model to increase the probability of target token $y_t$ while decreasing that of foil token $y_f$.
A foil token can be defined as an output with an alternative answer, suffering from incorrect generalization (\cref{sec:testbeds}).
Contrastive versions of \textsc{IE}, \textsc{GN}, and \textsc{I$\times$G} are defined as follows:
\begin{align}
    &\!\! S_{\mathrm{IE}}^*(x_i, y_t, y_f; \bm X) 
    \nonumber
    \\
    &\!\! \qquad = S_{\mathrm{IE}}(x_i, y_t; \bm X) - S_{\mathrm{IE}}(x_i, y_f; \bm X)
\end{align}
\begin{align}
    &\!\! S_{\mathrm{GN}}^*(\bm x_i, y_t, y_f; \bm X) = \lVert g^*(\bm x_i, y_t, y_f; \bm X) \rVert_{\mathrm{L1}}
    \\
    &\!\! S_{\mathrm{I\times G}}^*(\bm x_i, y_t, y_f; \bm X) \! = \! g^*(\bm x_i, y_t, y_f; \bm X) \cdot \bm x_i
    \\
    &\!\! g^*(\bm x_i, y_t, y_f; \bm X) = \nabla_{\bm x_i}\bigl( q(y_t \vert \bm X) - q(y_f \vert \bm X) \bigr)
\end{align}
\noindent

\subsection{Interpreting in-context learning (ICL)}
\label{sec:prompt-format}

We focus on the ICL setting~\cite{brown-etal-2020-learners}, which has typically been adopted in modern LLM-based reasoning.
An input prompt in ICL setting consists of few-shot examples $E$ and a target question.
$E$ is composed of $n$ examples [$e_1, \cdots, e_n$], each of which contains an input-output pair $e_i=(X_i, f(X_i))$, given a function $f$ associated with the task. 
Let $X_{n+1}$ represent the target question $q$ that the model must answer. 
The ICL setting is denoted as follows:
\begin{align*}
\underbrace{\overbrace{\overbrace{X_1, f(X_1)}^{\textit{example}~e_1}, \ldots, \overbrace{X_{n}, f(X_{n})}^{\textit{example}~e_n}}^{\textit{few-shot examples E}}, \overbrace{X_{n+1}}^q}_{\textit{prompt}\ X}, \underbrace{f(X_{n+1})}_{\textit{completion\ y}}
\end{align*}
\noindent
Here, a model is expected to first induce the underlying function (rule) $f$ from examples $E$ and then generate the final output $f(X_{n+1})$.

\paragraph{Aha example}
Interpreting a model's in-context learning (ICL) involves %
identifying when, within the input, the model infers the correct rule $f$. 
To address this aspect, we propose a unique benchmark that features an explicit ``aha moment'' ($e^*$) within the input prompt. At this moment, the correct rule $f$ can be identified by comparing the aha example with one of the other examples in the prompt. 
Thus, at least, $e^*$ should be one of the two most important examples (see evaluation metrics in~\cref{subsec:metrics}).
Note that, to mitigate the potential confusion, we exclude the case of $e^*$ being the first example in the demonstration since, in this case, its next example $e_2$ can disambiguate the rule and virtually work as the aha example from the perspective of the incremental reasoning process.

\paragraph{Instance-level attribution}
Notably, we consider the use of IA methods to identify a particular example $e^* \in E$ in input, instead of a particular token. %
To compute an IA score for an example $S(e_i)$, we sum up the IA scores for its constituent tokens\footnote{See Appendix~\ref{appendix:max-main-result} for results obtained with \emph{max} aggregation instead of sum; see Appendix~\ref{appendix:coalition-result} for results that treat the tokens in each example as a coalition and compute their attribution.}: $S(e_i) = \sum_{x_j \in (X_i, y_i) = e_i} S(x_j)$.\footnote{An exception applies in the IE method; the attribution score for an example $e_i$ is simply computed by erasing the corresponding $X_i$ and $f(X_i)$ from the input sequence.}
Our interest is which example obtains a high IA score.

\begin{table*}[t]
\small
\centering
\footnotesize
\tabcolsep 2pt
\begin{tabular}{llll}
    \toprule
    Task & Prompt example/template & Answer & Potential rules \\
    \cmidrule(r){1-1} \cmidrule(r){2-2} \cmidrule(r){3-3} \cmidrule(r){4-4}
    \textsc{Linear-or-Distinct} & 
    \begin{tabular}{l}
        \texttt{a a b a $\mapsto$ b}\\
        \texttt{g g j g $\mapsto$ j}\\
        \texttt{k i k k $\mapsto$ \textcolor[HTML]{4485F5}{k} / \textcolor[HTML]{D97757}{i}} \\
        \texttt{o o o p $\mapsto$}
    \end{tabular} & 
    \texttt{\textcolor[HTML]{4485F5}{o} / \textcolor[HTML]{D97757}{p}} & 
    \begin{tabular}{l}
        A. Generate the \textcolor[HTML]{4485F5}{\textit{n}-th token} \\ $\;\;\;\;$ (3rd token in this example)\\
        B. Generate the  \textcolor[HTML]{D97757}{distinctive token}\\
    \end{tabular} \\
    \cmidrule(r){1-1} \cmidrule(lr){2-2} \cmidrule(lr){3-3} \cmidrule(lr){4-4}
    \textsc{Add-or-Multiply} & 
    \begin{tabular}{l}
        \texttt{aa $\mapsto$ aaaa} \\
        \texttt{hh $\mapsto$ hhhh} \\
        \texttt{vvv $\mapsto$ \textcolor[HTML]{4485F5}{vvvvv} / \textcolor[HTML]{D97757}{vvvvvv}} \\
        \texttt{i $\mapsto$}
    \end{tabular} & 
    \texttt{\textcolor[HTML]{4485F5}{iii} / \textcolor[HTML]{D97757}{ii}} & 
    \begin{tabular}{l}
        A. \textcolor[HTML]{4485F5}{Add} $m$ tokens \\
        $\;\;\;\;$ ($m=2$ in this example) \\
        B. \textcolor[HTML]{D97757}{Multiply} the numder of tokens by $n$ \\
        $\;\;\;\;$ ($n=2$ in this example) \\
    \end{tabular} \\
    \cmidrule(r){1-1} \cmidrule(lr){2-2} \cmidrule(lr){3-3} \cmidrule(lr){4-4}
    \textsc{Verb-Object} & 
    \begin{tabular}{l}
        \textit{like} \texttt{[CITY]} $\mapsto$ True \\
        \textit{love} \texttt{[ANIMAL]} $\mapsto$ False\\
        \textit{like} \texttt{[ANIMAL]} $\mapsto$ \textcolor[HTML]{4485F5}{True} / \textcolor[HTML]{D97757}{False}\\
        \textit{love} \texttt{[CITY]} $\mapsto$
    \end{tabular} &
    \textcolor[HTML]{4485F5}{False} / \textcolor[HTML]{D97757}{True} & 
    \begin{tabular}{l}
        A. If \textcolor[HTML]{4485F5}{``like''} exists, then True\\
        B. If \textcolor[HTML]{D97757}{[CITY]} exists, then True \\
        
    \end{tabular} \\
    \cmidrule(r){1-1} \cmidrule(lr){2-2} \cmidrule(lr){3-3} \cmidrule(lr){4-4}
        \textsc{Tense-Article} & 
    \begin{tabular}{l}
        \textit{The} \texttt{[NOUN]} \texttt{[VERB]-ing} $\mapsto$ True \\
        \textit{A} \texttt{[NOUN]} \texttt{[VERB]-past} $\mapsto$ False\\
        \textit{A} \texttt{[NOUN]} \texttt{[VERB]-ing} $\mapsto$ \textcolor[HTML]{4485F5}{True} / \textcolor[HTML]{D97757}{False}\\
        \textit{The} \texttt{[NOUN]} \texttt{[VERB]-past} $\mapsto$
    \end{tabular} & 
    \textcolor[HTML]{4485F5}{False} / \textcolor[HTML]{D97757}{True} & 
    \begin{tabular}{l}
        A. If the verb is in \textcolor[HTML]{4485F5}{ing} form, then True\\
        B. If the first token is \textcolor[HTML]{D97757}{``the"}, then True \\
    \end{tabular} \\
    \cmidrule(r){1-1} \cmidrule(lr){2-2} \cmidrule(lr){3-3} \cmidrule(lr){4-4}
        \textsc{Pos-Title} & 
    \begin{tabular}{l}
        The \texttt{[NOUN]} Was \texttt{[ADJ]} $\mapsto$ True \\
        The \texttt{[noun]} was \texttt{[noun]}  $\mapsto$ False\\
        The \texttt{[noun]} was \texttt{[adj]} $\mapsto$ \textcolor[HTML]{4485F5}{True} / \textcolor[HTML]{D97757}{False}\\
        The \texttt{[NOUN]} Was \texttt{[NOUN]} $\mapsto$
    \end{tabular} & 
    \textcolor[HTML]{4485F5}{False} / \textcolor[HTML]{D97757}{True} & 
    \begin{tabular}{l}
        A. If \textcolor[HTML]{4485F5}{adjective} exsist, then True\\
        B. If the sentence is in \textcolor[HTML]{D97757}{title case}, then True \\
    \end{tabular} \\
    \midrule
    \midrule
    \textsc{Associative-Recall} & 
    \begin{tabular}{l}
        \texttt{a $\mapsto$ 6} \\
        \texttt{g $\mapsto$ 3}\\
        \texttt{w $\mapsto$ 5}\\
        \texttt{g $\mapsto$}
    \end{tabular} & 
    3 & 
    \begin{tabular}{l}
        Key--value pairs are in the prompt. \\ The task is to output a value associated \\ with a given key. 
    \end{tabular} \\

    \bottomrule
\end{tabular}
\caption{Formats of our inductive reasoning tasks. As a baseline setting, we also set \textsc{Associative-Recall} setting to just memorize key-value mappings. The remaining tasks span from somewhat superficial features to linguistic ones. The disambiguating example (the third one in these examples) determines the correct rule and answer (\textcolor[HTML]{4485F5}{blue} or \textcolor[HTML]{D97757}{orange}) for the final question from two plausible generalizations shown in the ``Potential rules'' column.}
\label{tab:task_description}

\end{table*}

\section{Problem settings}
\label{sec:testbeds}
We evaluate the performance of each IA method in identifying the crucial in-context example $e^*$ necessary for defining the task. 
In real-world tasks, it is generally unclear which in-context example is the most influential in solving the task, and the task may be solved even without relying on any of the examples (e.g., solved by leveraging prior knowledge). 
Therefore, these are not suitable as a benchmark to evaluate the interpretability method, and we design a synthetic and controlled tasks.

Our setting is the extension of \citet{Mueller2024-ae}; we employ a set of ambiguous inductive learning scenarios inspired by the cognitively-motivated LM analyses (\citealt{mccoy-etal-2020-syntax},\citealt{Warstadt2020-wo}; \textit{inter alia}).
In these scenarios, a task $f$ is mostly ambiguous in demonstrations $E$ in the sense that several compatible rules exist to explain the transformations $X\mapsto f(X)$.
We extend this setting by adding only one disambiguating example $e^*$ (``\textit{aha example}''), which determines the correct rule $f^*$ to be unique, and test whether each IA method can identify this special example as long as models correctly employ this clue $e^*$ to resolve the problem.
For instance, most examples shown in Figure~\ref{fig:fig1} are ambiguous (with gray color) w.r.t. the two possible rules of (i) adding the same token twice or (ii) multiplying the number of tokens by two.
This ambiguity is resolved by comparing the aha example $e^*$ (blue example in Figure~\ref{fig:fig1}) with any one of the other ambiguous examples. 
As shown in Table~\ref{tab:task_description}, we designed the following tasks as a case study:

\paragraph{\textsc{Linear-or-Distinct~(LD)}}
The few-shot examples are ambiguous as to Rule A: selecting a character in a particular linear position in an input $X_i$; or Rule B: selecting a character that differs from the others in an input $X_i$.

\paragraph{\textsc{Add-or-Multiply~(AM)}}
The ambiguity of this task is Rule A: add a certain number of tokens to input $X_i$; or Rule B: multiply the numbers of tokens in the input $X_i$.

\paragraph{\textsc{Verb-Obeject~(VO)}}
This task requires distinguishing whether the type of verb (Rule A) or the category of the object noun~(Rule B) matters.
We employed two verbs~(``like'' and ``love'') and two categories of the object~(\texttt{city} or \texttt{animal}).

\paragraph{\textsc{Tense-Article~(TA)}}
The potential rules are Rule A: whether the main verb in the input $X_i$ is in ing-form or not; or Rule B: whether the first token of input $X_i$ is ``The'' or not.

\paragraph{\textsc{Pos-Title~(PT)}}
This task involves two rules: %
Rule A: whether there is an adjective in $X_i$; or Rule B: whether $X_i$ is presented in the title case.
\begin{figure*}[t]
\centering
\includegraphics[width=\linewidth]{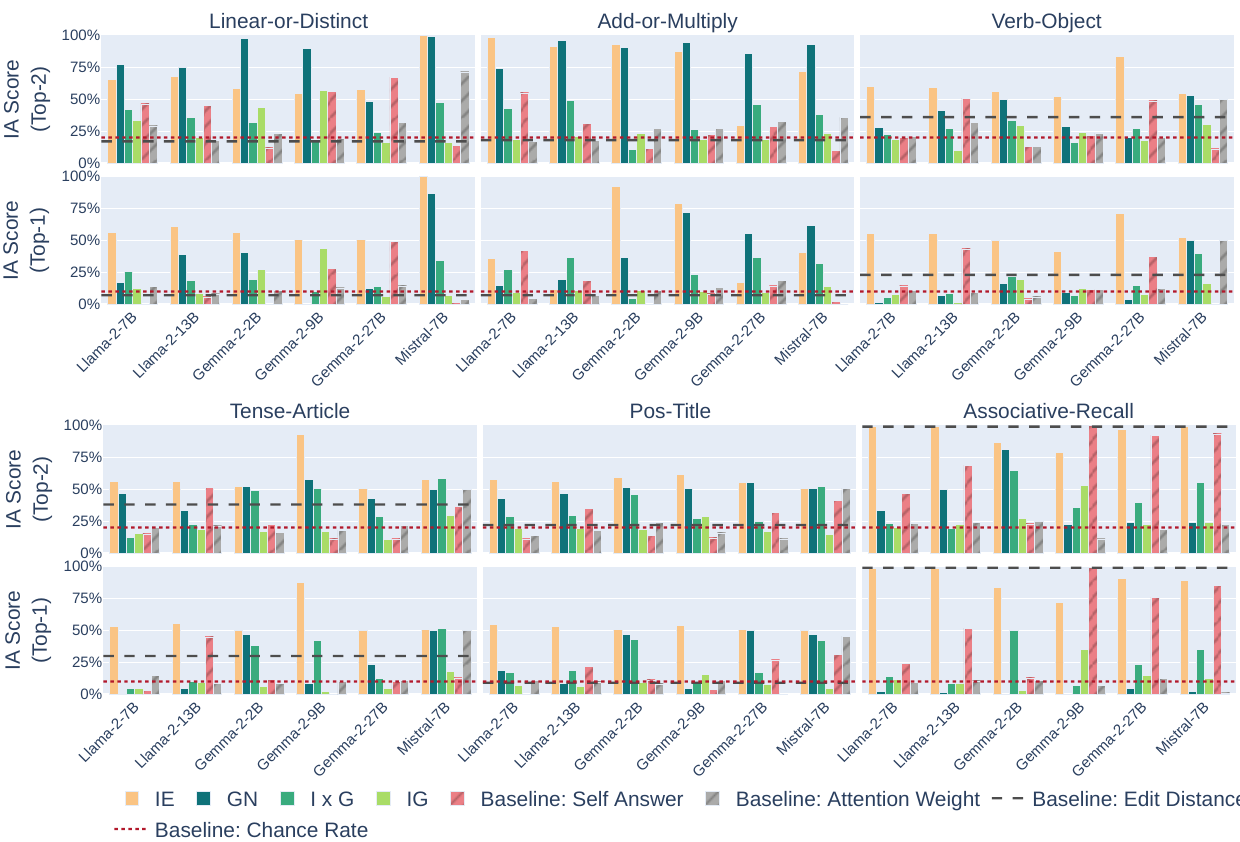}
\caption{IA scores (attribution accuracies) for each task/model in the 10-shot setting (thus, the chance rate is 20\% and 10\% for the top-two and top-one metric, respectively; red dotted line). The edit distance and attention baselines are indicated by a black dotted line and gray bar, respectively.}
\label{fig:main_result}
\end{figure*}
\\

In addition to them, we adopted a simple task of associative recall (AR), which is typically employed in studying ICL, where the model is supposed to simply memorize the key: value mapping rules demonstrated in the prompt and apply them to the target question.
Linguists may be more interested in the task of, for example, syntactic transformation to an interrogative sentence~\cite{mccoy-etal-2020-syntax} based on the original poverty of the stimulus argument in the language domain~\cite{chomsky1980rules}. However, the models' inference on such a realistic task may interfere with their meta-linguistic knowledge, which is inherently hard to track with IA methods; thus, we adopted synthetic tasks.\footnote{See Appendix~\ref{appendix:cot-setting} for results on tasks with chain-of-thought format.}

\paragraph{Foil token}
A contrastive explanation needs a foil token corresponding to an explicit negative label (\cref{sec:method-def}).
We use the token/answer %
corresponding to an alternative rule (conflicting the disambiguating example) as the foil token.

\section{Experimental setup}
\subsection{Overview}
\paragraph{Few-shot settings}
We conducted experiments with different numbers of few-shot examples; specifically, we examined 10-shot, 50-shot, and 100-shot settings to test the robustness of IA methods toward somewhat longer demonstrations.

\paragraph{Data}
For each synthetic task, we create 360 different questions with different sets of few-shot examples and a target question.
In the \textsc{LD}, \textsc{AM}, \textsc{VO}, \textsc{TA}, and \textsc{PT} tasks, the correct rule is selected out of the two candidates (rules A or B shown in Table~\ref{tab:task_description}) in a 1:1 ratio. %
The position of the disambiguating example $e^*$ is randomly selected according to a uniform distribution over all positions except the first.

\paragraph{Models}
We evaluate six LLMs: Llama-2-7B, Llama-2-13B~\cite{touvron-etal-2023-llama2}, Gemma-2-2B, Gemma-2-9B, Gemma-2-27B~\cite{riviere-etal-2024-gemma-2}, and Mistral-7B~\cite{jiang2023mistral7b}.
As a prerequisite for our experiments, the models should be able to learn and solve the task, i.e., be sufficiently sensitive to the disambiguating example $e^*$ and use this to determine the correct rule.
To ensure this ability, we fine-tune these models on each task (see Appendix~\ref{appendix:ft-detail}), but the conclusions overall did not alter before and after fine-tuning (see Appendix~\ref{appendix:base-mosel-result}).
Furthermore, we excluded the instances where models yielded an incorrect answer, as interpreting the model's failure is beyond our scope.

\subsection{Metrics}
\label{subsec:metrics}
We report two accuracy measures: (i) $e^*$ is in the top two examples with the highest IA score (top-2 accuracy), and (ii) $e^*$ gets the highest IA score among the input examples (top-1 accuracy).
Top-2 accuracy is motivated by the fact that models should at least consider the $e^*$ plus any other example to identify the correct rule (as described in \cref{sec:prompt-format}).
Top-1 accuracy is motivated from a leave-one-out perspective; excluding the $e^*$ significantly hurts the task answerability, while excluding the other examples does not hurt the task ambiguity/complexity.

\subsection{Baselines}
\label{subsec:baselines}
Along with the IA methods introduced in~\cref{sec:method-def}, we evaluate four baseline methods.

\paragraph{Edit distance}
This method identifies $e^*$ simply using edit distance between the target example $X_{n+1}\oplus y_{n+1}$ and each example $X_i\oplus y_i$, where $\oplus$ is a string concatenation. 
Example with the minimum edit distance, thus the most similar example to the target question, is selected as an explanation. 
The weak performance of this problem probes that our experimental setting is so challenging that just relying on surface features does not resolve it.

\paragraph{Attention weights}
This method leverages attention weights, computed as the sum of attention weights across all tokens in input $X$. 
While attention weights are generally considered unreliable for model interpretation, we include this baseline to compare whether IA methods achieve superior performance.

\paragraph{Self-answer}
We also examine directly asking the models to generate their rationale. Specifically, we have models generate the most informative example in a prompt (Appendix~\ref{appendix:exact-prompt}) in deriving their answer as a post-hoc explanation.
This might be, more or less, relevant to the verbalization of \textit{aha moment} recently observed in DeepSeek models~\cite{guo2025deepseek}.

\paragraph{Chance rate}
We also report the chance rate of attribution accuracy when randomly selecting one example from a prompt.

\section{Experiments}
\label{sec:experiment}
Figure~\ref{fig:main_result} shows the results in 10-shot settings, with both top-2 and top-1 metrics.
Additional analyses are presented in Appendices~\ref{appendix:max-main-result},~\ref{appendix:cot-setting}, and \ref{appendix:position-bias}.
\subsection{Main results}
\label{ssec:main_results}

\paragraph{IE works the best}
First of all, the input erasure method generally performed the best in both top-1 and top-2 accuracies.
This is somewhat obvious because our task is designed to be unsolvable by removing the aha example and thus rather serves as a quick check for our experimental design.
Having said that, the input erasure method has some disadvantages in regard to the computational costs of repeated decoding by removing examples one by one as well as the unclarity of by which unit an erasure should be applied, especially in a real, somewhat noisy input.
Additionally, the accuracy was not 100\% in almost all the cases; we further discuss the potential flaw of this approach in \cref{sec:discussion}.

\paragraph{Potential of gradient-based approaches}
As for baselines, while the self-answer approach worked well in specific settings (associative recall with larger models), most baselines, including attention weights, generally failed to achieve high accuracy.
Edit distance was a somewhat strong baseline, but it has obvious limitations of lacking semantic similarities and was frequently outperformed by GN.
Compared to such baselines, the gradient-based method worked relatively well, highlighting the potential of this direction.

\paragraph{Improved versions of gradient-based methods do not outperform GN}
Among the gradient-based methods, simple gradient norm tends to work the best in most tasks, especially in top-2 accuracy. %
In other words, whereas I$\times$G and IG are proposed as the improved version of simple gradient norm method, there were no substantial advantages of these methods in our settings.
In particular, IG consistently yielded the lowest attribution accuracy across all six tasks among the gradient-based methods, suggesting its limitations in ICL scenarios.
The plausible reason behind this inferiority is discussed in~\cref{sec:discussion}.

\paragraph{General failure}
Nevertheless, some simple tasks, such as \textsc{Verb-Object}, \textsc{Tense-Article}, and \textsc{Pos-Title}, were ever hard to interpret with any approach. This opens a new field for developing a better interpretation method for ICL.

\subsection{Scaling properties}
\label{ssec:scaling}
In the age of LLMs, the setting has progressively been scaled up toward model parameter size and context length.
We analyze how such a scaling affects the LLMs' interpretability.

\paragraph{Interpretability vs. model size}
We first investigate the relationship between attribution accuracy and model size --- is it more difficult to interpret larger models?
We observe somewhat intriguing patterns for this question (Figure~\ref{fig:main_result}); gradient-based methods tend to work worse in larger models, and in contrast, the self-answer baseline works better in larger models (especially in \textsc{Linear-or-Distinct} and \textsc{Associative-Recall}).
That is, the (empirically) accurate approach to interpreting the LLMs may differ in their model scale, and the success in interpreting smaller models does not always entail the success in interpreting larger models.

\begin{figure*}[t]
    \centering
    \includegraphics[width=\linewidth]{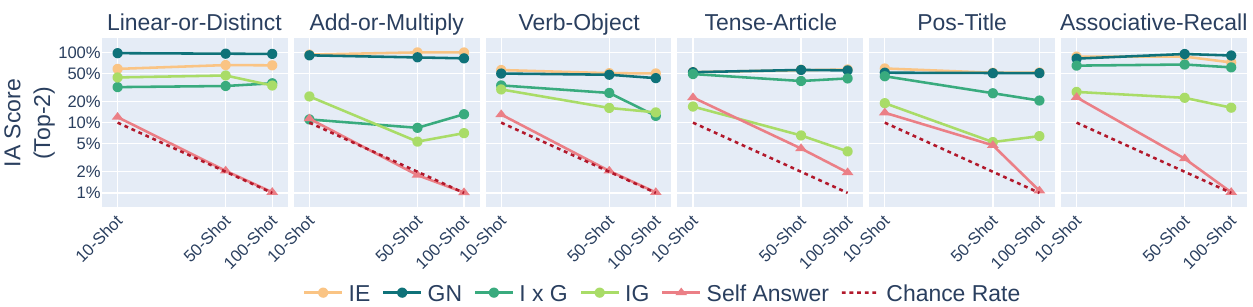}
    \caption{IA score (attribution accuracy) for interpreting Gemma-2-2B models across all six tasks. Gradient-based methods are relatively robust to the number of few-shot examples, while there is a consistent, large drop in attribution accuracy in SA. Note that both x-axis and y-axis are in log scale.}
    \label{fig:shot_number_scaling}
\end{figure*}

\paragraph{Interpretability vs. number of examples}
Next, given the trend of long-context LLMs, we examine the relationship between attribution accuracy and the number of few-shot examples.
Figure~\ref{fig:shot_number_scaling} shows the attribution accuracy for Gemma-2-2B in all six tasks in different numbers of in-context examples.
This demonstrates that gradient-based methods maintain accuracy or rather improve against the longer context, in contrast to the decreasing chance rate.
This suggests the robustness of IA methods in long-context scenarios, highlighting their potential for interpreting inputs with extensive contextual information.
Notably, the quality of self-answer consistently degraded as the number of in-context examples increased, while the positive scaling effect was observed in the previous analysis of model size. 
That is, the gradient-based methods and self-answer approach exhibit an insightful trade-off between different scaling properties.

\begin{table}[t]
\centering
\scriptsize

\begin{tabular}{crrr}
    \toprule
    \multirow{2}{*}{Task} & \multirow{2}{*}{Accuracy~(\%)} & \multicolumn{2}{c}{IE Attr. Acc. (\%)} \\
    \cmidrule(lr){3-4}
                         &                             & Top-2          & Top-1 \\
    \cmidrule(r){1-1} \cmidrule(lr){2-2} \cmidrule(lr){3-3} \cmidrule(lr){4-4}
    \textsc{LD}~(Rule A)      & $98.0$                      & \multirow{2}{*}{$58.2$} & \multirow{2}{*}{$56.0$} \\
    \textsc{LD}~(Rule B)      & $0.5$                       &                  &  \\
    \cmidrule(r){1-1} \cmidrule(lr){2-2} \cmidrule(lr){3-3} \cmidrule(lr){4-4}
    \textsc{AM}~(Rule A)      & $34.5$                      & \multirow{2}{*}{$93.1$} & \multirow{2}{*}{$92.2$} \\
    \textsc{AM}~(Rule B)      & $65.5$                      &                  &  \\
    \cmidrule(r){1-1} \cmidrule(lr){2-2} \cmidrule(lr){3-3} \cmidrule(lr){4-4}
    \textsc{VO}~(Rule A)      & $100.0$                     & \multirow{2}{*}{$56.1$} & \multirow{2}{*}{$50.3$} \\
    \textsc{VO}~(Rule B)      & $0.0$                       &                  &  \\
    \cmidrule(r){1-1} \cmidrule(lr){2-2} \cmidrule(lr){3-3} \cmidrule(lr){4-4}
    \textsc{TA}~(Rule A)      & $98.0$                      & \multirow{2}{*}{$52.5$} & \multirow{2}{*}{$50.0$} \\
    \textsc{TA}~(Rule B)      & $2.0$                       &                  &  \\
    \cmidrule(r){1-1} \cmidrule(lr){2-2} \cmidrule(lr){3-3} \cmidrule(lr){4-4}
    \textsc{PT}~(Rule A)      & $68.5$                      & \multirow{2}{*}{$59.0$} & \multirow{2}{*}{$51.1$} \\
    \textsc{PT}~(Rule B)      & $31.5$        &                  &  \\
    \bottomrule
\end{tabular}

\caption{Task accuracy (not attribution accuracy) of Gemma-2-2B (excluding \textsc{AR}) when the disambiguating example is not included, separated by the correct rule. The accuracy drastically differs when the correct rule is different; thus, the models adopt a particular default rule with their inductive biases against fully ambiguous demonstrations, even in our controlled settings.}
\label{tab:ie_discussion}
\end{table}

\section{Discussion}
\label{sec:discussion}
This section discusses the potential reasons for the unexpected results presented in \cref{sec:experiment}, highlighting the challenging issues in interpreting ICL.

\paragraph{Why did IE fail to achieve 100\% attribution accuracy?}

Our tasks can not be answered without disambiguating \textit{aha} examples. 
Thus, it is somewhat unintuitive to see the non-100\% attribution accuracy of the IE method (again, the LLMs understood the task as they achieved 100\% accuracy in the tasks) --- what happened here?
To obtain a hint to clarify IE's potential limitations, we analyze model behaviors when the disambiguating example is excluded.
Interestingly, LLMs adopted a specific generalization (rule) in each task when there was no disambiguating example (Table~\ref{tab:ie_discussion}); in other words, they sometimes exhibited strong inductive biases in our tasks.
That is, when the correct rule is equal to %
their preferred rule by their inductive bias, they can answer the task correctly even without disambiguating examples, and the IE method does not compute a proper attribution score.
It is now common to see LLMs have particular inductive biases (not a \textit{tabula rasa})%
~\cite{warstadt-etal-2020-learning, kharitonov-chaabouni-2021-what}. %
Catching such generalization bias with IA methods represents an inherent challenge, highlighting their potential limitations in interpreting LLMs.

\paragraph{Why was I$\times$G worse than GN?}
The advantage of the I$\times$G method, compared to the GN, is the consideration of the norm of the input embedding~\cite{shrikumar-etal-2017-learning, denil-etal-2014-extraction}. %
Since a large vector tends to have a large dot product with another vector, the norm of the input token embedding (vector) is expected to affect the IA score of I$\times$G.
Then, no improvement of the I$\times$G over the GN suggests that, at least in our settings, the norm of the embeddings was not informative to estimate the input attribution.
The norm of the embedding largely has decontextualized information about the word, such as frequency, and it may make sense that such information is not helpful to interpreting our controlled, synthetic ICL tasks consisting of alphabet characters, numbers, or random words.

\paragraph{Why was IG worse than GN?}
IG is a path-based approach; the gradient is accumulated from a baseline vector (typically a zero vector) to the targeted input representation (in our case, the sequence of input embeddings representing few-shot examples).
This approach is somewhat intuitive when considering an attribution for a particular word or sentence; for example,  suppose one computes an attribution to the word ``excellent'' in an input, IG may trace the path from zero to the ``excellent'' vector, which will go through the \textit{goodness} direction, involving the intermediate points corresponding to, e.g., ``okay'' ``decent,'' ``good,'' ``excellent''~\cite{Sanyal2021-si}.
Then, one critical question is --- what does this path mean in interpreting the prompt and task representations?
Different prompt representations will no longer correspond to the same targeted task; thus, the attribution of a particular token under the intermediate point to the target prompt vectors may no longer be an attribution under a targeted task.
This can be one concern regarding the ineffectiveness of IG in our settings.

\begin{figure}[t]
    \centering
    \includegraphics[width=\linewidth]{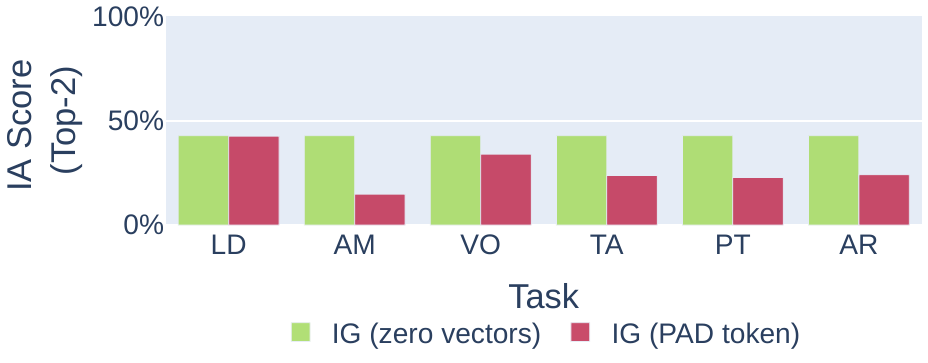}
    \caption{IA score (attribution accuracy) obtained with different baselines for IG when interpreting the Gemma-2-2B models across all six tasks. Zero vectors consistently yield higher accuracy than \texttt{<pad>} tokens.}
    \label{fig:alter_ig_baseline}
\end{figure}

\begin{figure*}[t]
    \centering
    \includegraphics[width=\linewidth]{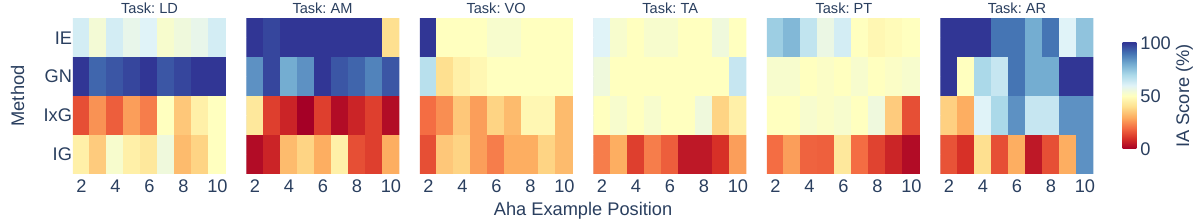}
    \caption{IA score (attribution accuracy) for different disambiguating, aha example positions. Gemma-2-2B model is used. All methods exhibit some degree of positional bias toward the aha example’s position.}
    \label{fig:g2b_position_bias}
\end{figure*}

\paragraph{Will the Baseline Choice for IG Change the Results?}
Existing work on BERT~\cite{devlin-etal-2019-bert} shows that the \texttt{[MASK]} token is a better baseline for IG, whereas the \texttt{[PAD]} token and zero vectors yield no substantial difference in performance~\cite{bastings-etal-2022-will}. 
Although we adopted zero vectors as the baseline in our main experiments, given that causal LMs do not have mask tokens, we further discuss the impact of this baseline choice on our results.  
We conduct a case study on the Gemma-2-2B models, using pad tokens (\texttt{<pad>}) as an alternative baseline for IG. The results are shown in Figure~\ref{fig:alter_ig_baseline}. The IA scores obtained with zero vectors consistently surpass those obtained with \texttt{<pad>} tokens, suggesting that zero vectors are the better choice for interpreting causal LLMs.

\paragraph{Positional Impact of the Aha Example}
LLMs are known to exhibit positional bias to prioritize information at the beginning or the end of the input~\cite{liu-etal-2024-lost}. 
Given this concern, we also examine the positional bias underlying our interpretability results.
Figure~\ref{fig:g2b_position_bias} shows attribution accuracy by the position of disambiguating, aha example. 
IA methods exhibited positional bias, though the patterns varied across tasks and methods.
For instance, in the VO task, IE and GN performed better when the aha example appeared at the beginning, while in the AR task, IG performed better when it appeared at the end.
See Appendix~\ref{appendix:position-bias} for other models' results as well as for 50 and 100-shot settings.

\section{Related Work}

\paragraph{IA methods}
Several lines of research are conducted to interpret neural language models.
NLP researchers have adapted IA methods, which were originally applied to vision models~\cite{simonyan-etal-2014-saliency, springenberg-etal-2015-striving, zintgraf-etal-2017-visualizing}, to perform a post-hoc interpretation of input-output associations exploited by language models~\cite{karpathy-etal-2015-visualizing, li-etal-2016-visualizing, arras-etal-2016-explaining, lei-etal-2016-rationalizing, alvarez-melis-jaakkola-2017-causal}, and its improved versions have also been developed~\cite{denil-etal-2014-extraction, sundararajan-etal-2017-axiomatic,murdoch-etal-2018-beyond, sinha-etal-2021-perturbing, ding-koehn-2021-evaluating, bastings-etal-2022-will,yin-neubig-2022-interpreting,ferrando-etal-2023-explaining}.
In line with these studies, we provide a new perspective to evaluate these IA methods in ICL.
Note that, as an orthogonal attempt, some research estimates the saliency scores to directly prompt models to generate such explanations~\cite{rajani-etal-2019-explain, liu-etal-2019-towards-explainable, wu-mooney-2019-faithful, narang-etal-2020-wt5, marasovic-etal-2022-few-shot}.
This method is indeed examined as one baseline in our study.

\paragraph{Instance-based explanation}
Instance-based explanation seeks explanation in the training data rather than relying on immediate input during inference, as IA methods do~\cite{wachter-etal-2017-conterfactual, charpiat-etal-2019-input, hanawa-etal-2021-evaluation}.
These two paradigms of instance-based and IA-based explanations have been studied somewhat separately since the information source to seek the explanation is clearly different.
On the other hand, in ICL, the training examples are now in the input during inference that can be analyzed by IA methods. 
In this sense, our investigation can be seen as a new exploration of instance-based explanation with the help of IA methods.

\paragraph{Mechanistic Interpretability}
With the rise of large language models, such as GPT-3~\cite{brown-etal-2020-learners}, the mechanistic interpretability community has shifted its focus from vision models to language models. Within which, the promising results using sparse autoencoders (SAEs)~\cite{bricken2023monosemanticity, templeton2024scaling} have inspired a flurry of follow-up work~\cite{gao-etal-2024-scaling-autoencoders, liberum-etal-2024-gemma-scope, rajamanoharan2024improving, rajamanoharan-etal-2024-jumping, karvonen2024measuring, braun-etal-2024-identifying, kissane2024interpreting, makelov2024sparse}.
Such a scope of SAE, interpreting the model internals, is orthogonal to our direction of estimating the importance of input examples.

\section{Conclusions}

We have pointed out and tackled the problem of interpreting the inductive reasoning process in ICL as a missing but reasonable milestone to be explored in LLM interpretability research.
Our revisit to the IA methods in interpreting the ICL process has clarified their limitations from a new angle as well as provided fruitful insights and discussions on their practical usage in modern NLP.
Some methods frequently failed even on a very simple associative recall task, and there is room to sophisticate previously developed interpretability tools to be suitable for LLMs.

\clearpage
\section*{Limitations}
Our study has several limitations in scope. First, we focused primarily on popular gradient-based IA methods, leaving other approaches such as perturbation-based methods like LIME~\cite{ribeiro-etal-2016-trust} and SHAP~\cite{lundberg-and-lee-shap} for future work.
Furthermore, the optimization of the interpretation methods, e.g., the prompt used in the self-answer baseline, was not explored in depth.

Second, regarding model selection, we concentrated on widely-used open-weight LLMs. Since applying IA methods requires gradient computation through backward propagation, computational constraints limited our ability to evaluate all available models, particularly large ones such as Llama-2-70B~\cite{touvron-etal-2023-llama2}.
In addition, we focused exclusively on pre-trained models, excluding post-trained (e.g., instruction-tuned) models from our analysis, although this choice was motivated by starting from interpreting simpler few-shot abilities that emerged during the pre-training phase.

Third, our experimental design used synthetic tasks to better define influential examples in the few-shot setting. While this approach allowed for controlled experimentation, both the number and format of tasks were limited. The design of the synthetic dataset is also biased toward the input erasure method. Future work could explore more realistic tasks with greater variations.

Finally, we may have to acknowledge that models might interpret these tasks differently than intended. As described in the discussion, models may sometimes rely on their inductive biases that can not be attributed to input tokens. Tracing their rationale not only into input but also model internals may also be needed to fully interpret the model's inductive reasoning process.

\section*{Ethical Statements}
This work explores input attribution~(IA) methods over large language models' (LLMs) in-context learning~(ICL) ability. Our findings contribute to the broader goal of developing more interpretable and safer AI systems by providing practical insights into the strengths and weaknesses of IA methods as tools for interpreting LLMs.
This study exclusively uses synthetic data generated through computational methods. No real user data, human annotations, or personally identifiable information were collected or used in our experiments. Our synthetic dataset generation process did not involve any human subjects, crowd workers, or demographic information.
We employ LLMs for writing assistance, specifically for grammar checking and polishing the manuscript. All LLM outputs were subsequently reviewed and edited by the authors to ensure accuracy and fidelity to the intended message.

\section*{Acknowledgments}
We want to express our gratitude to the members of the Tohoku NLP Group for their insightful comments. And special thanks to Sho Yokoi for his valuable suggestions on how to improve clarity in several aspects. This work was supported by the JSPS KAKENHI Grant Number JP24H00727,
JP22J21492; JST Moonshot R\&D Grant Number JPMJMS2011-35 (fundamental research); JST BOOST, Japan Grant Number JPMJBS2421; Grant-in-Aid for Early-Career Scientists Grant Number JP23K16938.

\bibliography{anthology,custom}
\clearpage
\appendix
\section{Aggregating attritbuion with \textit{max}}
\label{appendix:max-main-result}
Figure~\ref{fig:max_main_result} presents the IA scores using maximum aggregation to convert token-level attribution to example-level attribution: $S(e_i) = \max_{x_j \in (X_i, y_i) = e_i} S(x_j)$.
The overall trend for all the IA methods is consistent with the sum aggregation (Figure~\ref{fig:main_result}); thus, our results can be generalized regardless of this design.

\section{Attribution with coalition}
\label{appendix:coalition-result}
Figure~\ref{fig:coalition} presents the IA scores obtained with coalition aggregation, that is, we treat each example as a single coalition by computing the mean pooling of all the token embeddings in the example to obtain its “coalition embedding”, and then calculate its attribution. Specifically, instead of using the embedding of each token $\mathbf{X} \coloneqq [x_1, \ldots, x_n]$, we use the mean-pooled embedding $\bar{X} = \frac{1}{n}\sum_{i=1}^{n} x_i$.
The difference between calculating token-level attributions and then aggregating them with the coalition approach is minimal, except for the IxG metric in the \textsc{Pos-Title} task.

\section{Finetuning details}
\label{appendix:ft-detail}
The fine-tuning dataset consisted of 400 tasks for each of the 10-shot, 50-shot, and 100-shot settings (1,200 tasks total). For each task, we created a training set for fine-tuning using tokens that did not overlap with our test set (the dataset used in our main experiments). We fine-tuned models separately on each task, resulting in six fine-tuned models per LLM. The exception was the Gemma-2-27B model, which we did not fine-tune on the \textsc{Associative-Recall} task since the original model already performed well enough on this simple task.

\subsection{Finetuing parameters}
\label{appendix:ft-parameters}
We use a consistent LoRA configuration with rank $r=32$ and scaling factor $\alpha=64$, applying a dropout rate of $0.05$ across all linear modules. The LoRA adaptation includes bias terms in the training.
For optimization, we perform a learning rate sweep using a cosine scheduler with a $5\%$ warm-up period relative to the total training steps. The optimal learning rate typically falls in the order of $1\times10^{-5}$ when the loss reaches its minimum. 
In our experiments, the Llama-2 models achieved nearly zero loss, which is expected in such a synthetic setting. The Gemma-2 models, however, converge to final loss values of approximately $0.2$.

\subsection{Zero-shot task accuracy after finetuning}
We evaluate task accuracy using exact match, with results presented in Tables~\ref{tab:finetuning_accuracy},~\ref{tab:g2_9b_zero_acc},~\ref{tab:g2_27b_zero_acc},~\ref{tab:l2_7b_zero_acc}, and~\ref{tab:l2_13b_zero_acc}. In the zero-shot setting, some tasks show accuracies significantly below chance rate~(indicated in parentheses), as models occasionally generate unexpected responses. Notably, all models achieve zero-shot accuracies at or below the chance rate across all tasks, suggesting that models cannot solve our tasks relying only on the aha example.

\label{appendix:zero-shot-acc}

\section{Base model results}
\label{appendix:base-mosel-result}
\begin{figure*}[t]
    \centering
    \includegraphics[width=\linewidth]{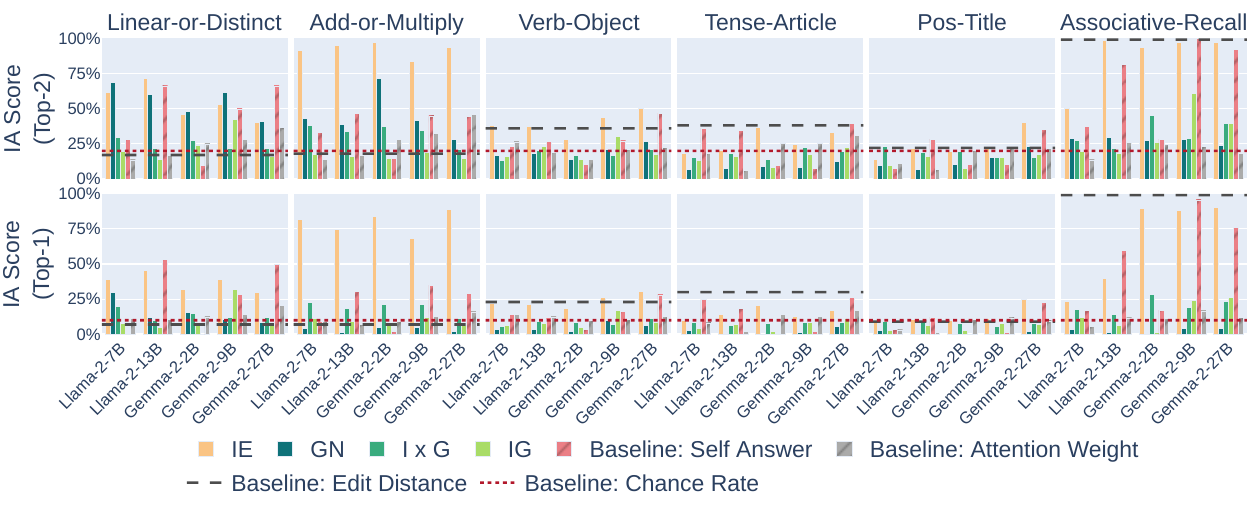}
    \caption{Attribution accuracies for each task for base models. Similar patterns to those observed in the fine-tuned models (Figure~\ref{fig:main_result}) can be obsered.}
    \label{fig:base_model_main_result}
\end{figure*}

\begin{figure*}[t]
    \centering
    \includegraphics[width=\linewidth]{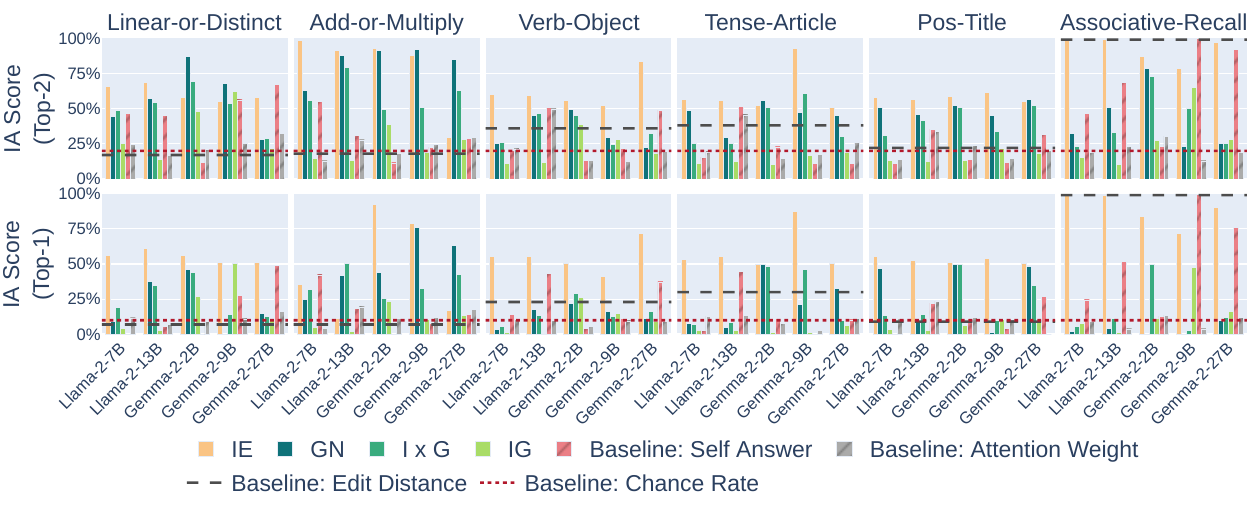}
    \caption{Attribution accuracies for each task use max aggregation. The overall trend for all IA methods is consistent with sum aggregation (Figure~\ref{fig:main_result})}
    \label{fig:max_main_result}
\end{figure*}

\begin{figure*}[t]
    \centering
    \includegraphics[width=\linewidth]{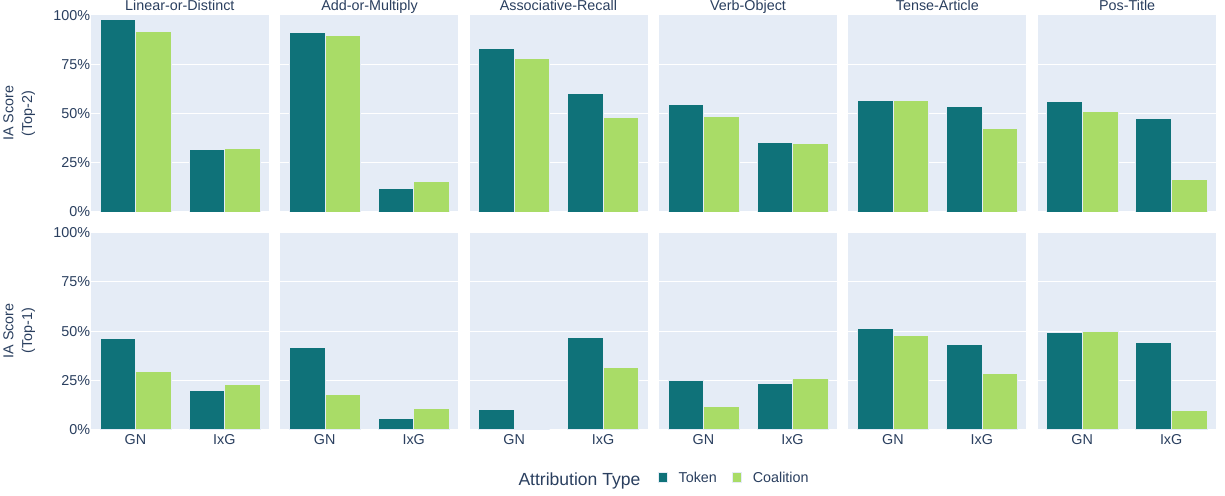}
    \caption{Coalition attribution accuracies for each task by treating each example as one coalition.}
    \label{fig:coalition}
\end{figure*}

Figure~\ref{fig:base_model_main_result} presents the IA scores for the base models. While the overall IA scores for \textsc{VO}, \textsc{TA} and \textsc{PT} tasks are relatively low, the performance trends across different tasks and models exhibit similar patterns to those observed in the fine-tuned models (Figure~\ref{fig:main_result}). Therefore, our results can be generalized regardless of fine-tuning.

\section{Prompts}
\label{appendix:exact-prompt}
We present sample of the exact prompt we used for our task, including the ones we used for testing attribution accurices and modified prompts for self-answer.
Note that in all the experiments, we only used the model outputs with a correct answer.
That is why we appended the correct answer in advance to the self-answer prompt to obtain the post-hoc explanation.

\paragraph{Normal Prompt}
\graybox{Input: they, Output: 6\\
    Input: not, Output: 3\\
    Input: I, Output: 5\\
    Input: tell, Output: 7\\
    Input: them, Output: 6\\
    Input: were, Output: 6\\
    Input: at, Output: 0\\
    Input: yes, Output: 1\\
    Input: right, Output: 9\\
    Input: say, Output: 3\\
    Input: they, Output:}

\paragraph{Self-answer Prompt}
\graybox{<0>Input: they, Output: 6</0>\\
<1>Input: not, Output: 3</1>\\
<2>Input: I, Output: 5</2>\\
<3>Input: tell, Output: 7</3>\\
<4>Input: them, Output: 6</4>\\
<5>Input: were, Output: 6</5>\\
<6>Input: at, Output: 0</6>\\
<7>Input: yes, Output: 1</7>\\
<8>Input: right, Output: 9</8>\\
<9>Input: say, Output: 3</9>\\
<target>Input: they, Output: </target>\\
\\
Among the 10 examples labeled <0> to <9>, select the single most helpful example for determining the answer to the <target> question. The correct answer to the target question is ``6''. To conclude this answer, we need to find one example that provides the necessary information. Therefore, the most helpful example is <}

\section{Chain-of-though format}
\label{appendix:cot-setting}
To contextualize our experimental settings with more practical scenarios, we further evaluate attribution accuracies on top of chain-of-thought (CoT) prompting~\cite{wei-etal-2022-cot}.
We use Gemma-2-27B-IT (a post-training version of Gemma-2-27B) instead of the base model to perform a better CoT-style generation and employ the \textsc{AM} task, where the model achieved high accuracy with CoT, as a case study.
We only target the last time step to generate the exact answer.
We compute the by-example attribution scores the same as our main experiments, but now the attribution scores can be spread over the reasoning chain part as well as in-context examples.
Our target is which example is informative to answer the question; thus, the attribution to the chain part is tentatively disregarded.
As statistics, we just report how many proportions of attribution scores ([0-100\%]) reached the reasoning chain part (denoted as ``Chain prop.'').

The results are presented in Table~\ref{tab:cot-task}. 
All tested IA methods performed worse in this CoT setting than in the CoT-free settings in the main experiments.
Nevertheless,  the superiority of GN to other approaches still holds. 
Note that the Chain prop. substantially differs across IA methods; for example, IE assigns over 80\% of the attribution score to the chain.
These divergent results also suggest that the conventional IA methods can not easily be applied to modern ICL and CoT settings.

The exact prompt and the reasoning chain generated by the model are provided below\footnote{The prompt template is applied since this is a post-training model}:
\paragraph{CoT Prompt}
\graybox{<start\_of\_turn> user\\
Input: saw, 2, Output: saw, 4\\
Input: start, 2, Output: start, 4\\
Input: the, 2, Output: the, 4\\
Input: too, 2, Output: too, 4\\
Input: round, 2, Output: round, 4\\
Input: which, 1, Output: which, 3\\
Input: work, 2, Output: work, 4\\
Input: get, 2, Output: get, 4\\
Input: that, 2, Output: that, 4\\
Input: white, 2, Output: white, 4\\
Input: I, 3, Output: <ANSWER>\\[1em]
Solve this problem step by step, generate the content of <ANSWER> after ``So the answer is'': <end\_of\_turn>\\[1em]
<start\_of\_turn> model}

\begin{table}[t]
\centering
\scriptsize

\begin{tabular}{lrrlr}
    \toprule
    Method & \multicolumn{2}{c}{IA score} & Aggregation & Chain prop.~(\%) \\
    \cmidrule(lr){2-3}
           & Top-1 & Top-2 &  &  \\
    \cmidrule(r){1-1} \cmidrule(l){2-3} \cmidrule(l){4-4} \cmidrule(l){5-5}
    IE           & $11.3$ & $17.4$ & \multirow{4}{*}{Sum} & $82.1$ \\
    GN           & $12.4$ & $37.6$ &                      & $35.4$ \\
    I $\times$ G &  $14.9$ & $29.8$ &                      & $23.0$ \\
    IG           & $8.9$ & $22.7$ &                      & $42.9$ \\
    \midrule
    \midrule
    IE           & $11.3$ & $17.4$ & \multirow{4}{*}{Max} & $82.1$ \\
    GN           & $14.5$ & $33.3$ &                      & $19.2$ \\
    I $\times$ G &  $9.9$ & $31.2$ &                      & $23.1$ \\
    IG           & $9.9$ & $22.3$ &                      & $7.5$ \\
    \bottomrule
\end{tabular}

\caption{IA score for the CoT-prompted \textsc{AM} task. The percentage of attribution scores allocated to the reasoning chain is denoted as Chain prop.}
\label{tab:cot-task}
\end{table}

\section{Distribution of example with the highest attribution score}
\label{appendix:position-bias}
\begin{figure*}[t]
    \centering
    \includegraphics[width=0.98\linewidth]{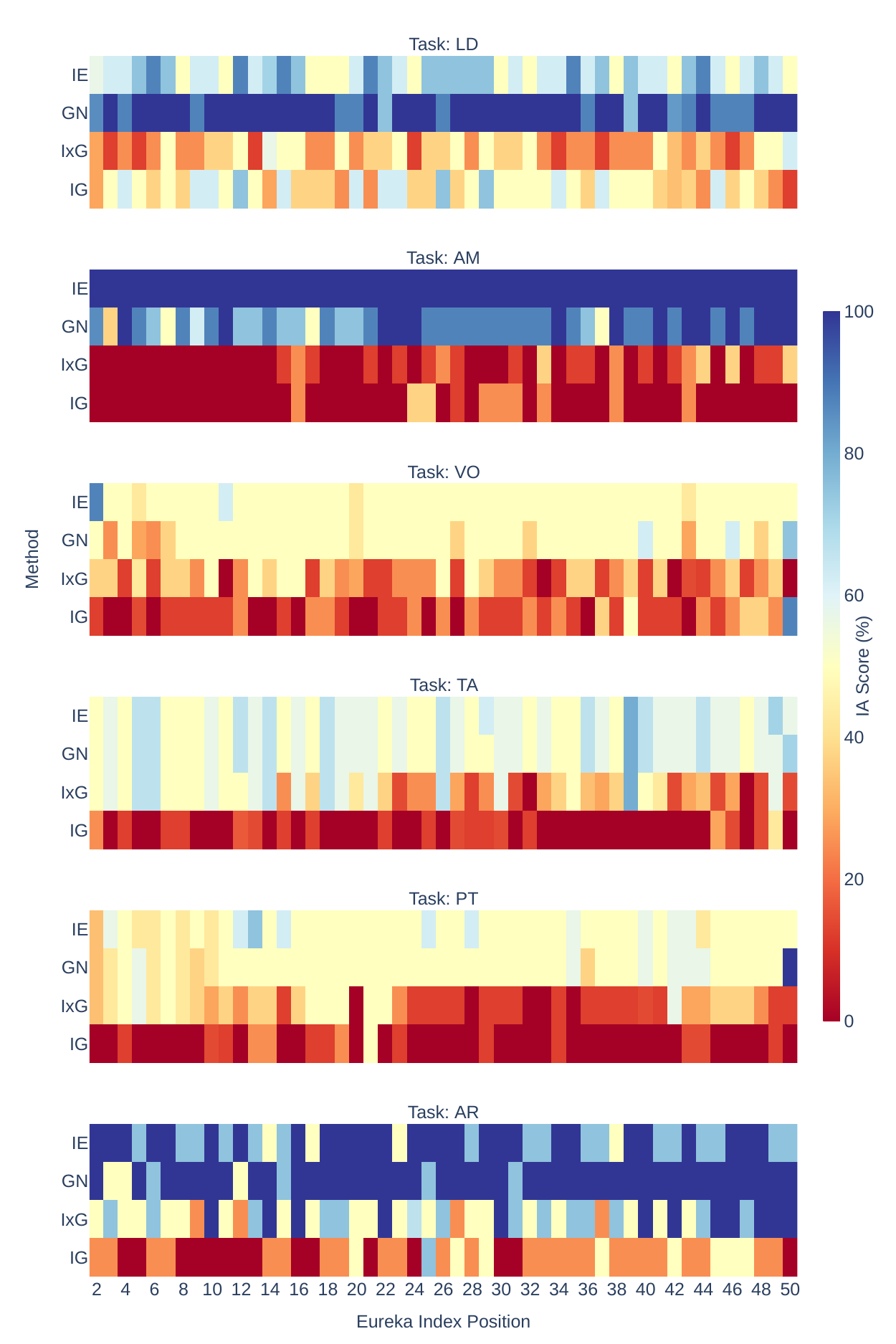}
    \caption{Distribution of the position of the example with the highest attribution scores across IA methods (Gemma-2-2B model), 50-Shot.}
    \label{fig:g2b_distribution_50}
\end{figure*}

\begin{figure*}[t]
    \centering
    \includegraphics[width=0.98\linewidth]{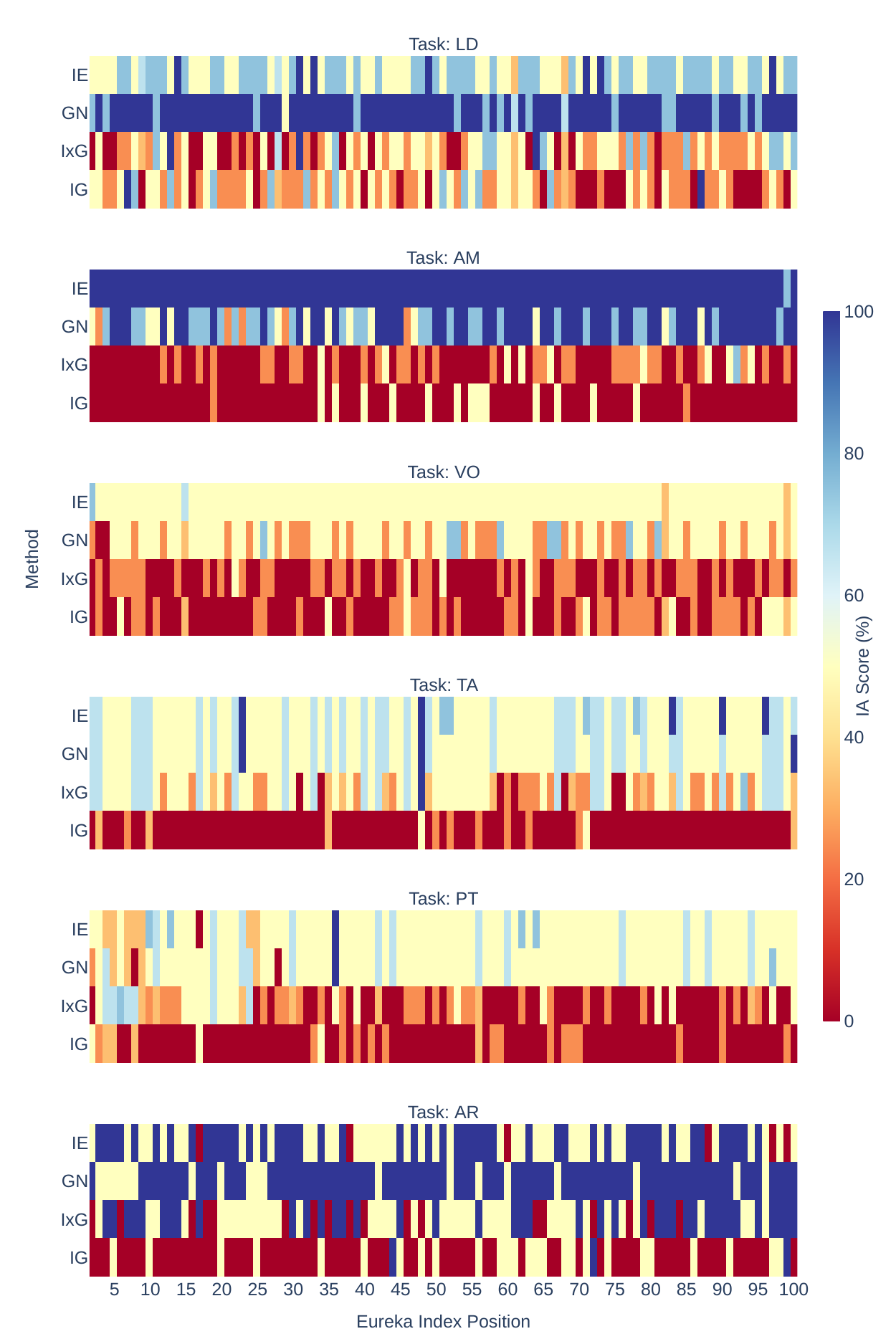}
    \caption{Distribution of the position of the example with the highest attribution scores across IA methods (Gemma-2-2B model, 100-Shot).}
    \label{fig:g2b_distribution_100}
\end{figure*}

\begin{figure*}[t]
    \centering
    \includegraphics[width=0.98\linewidth]{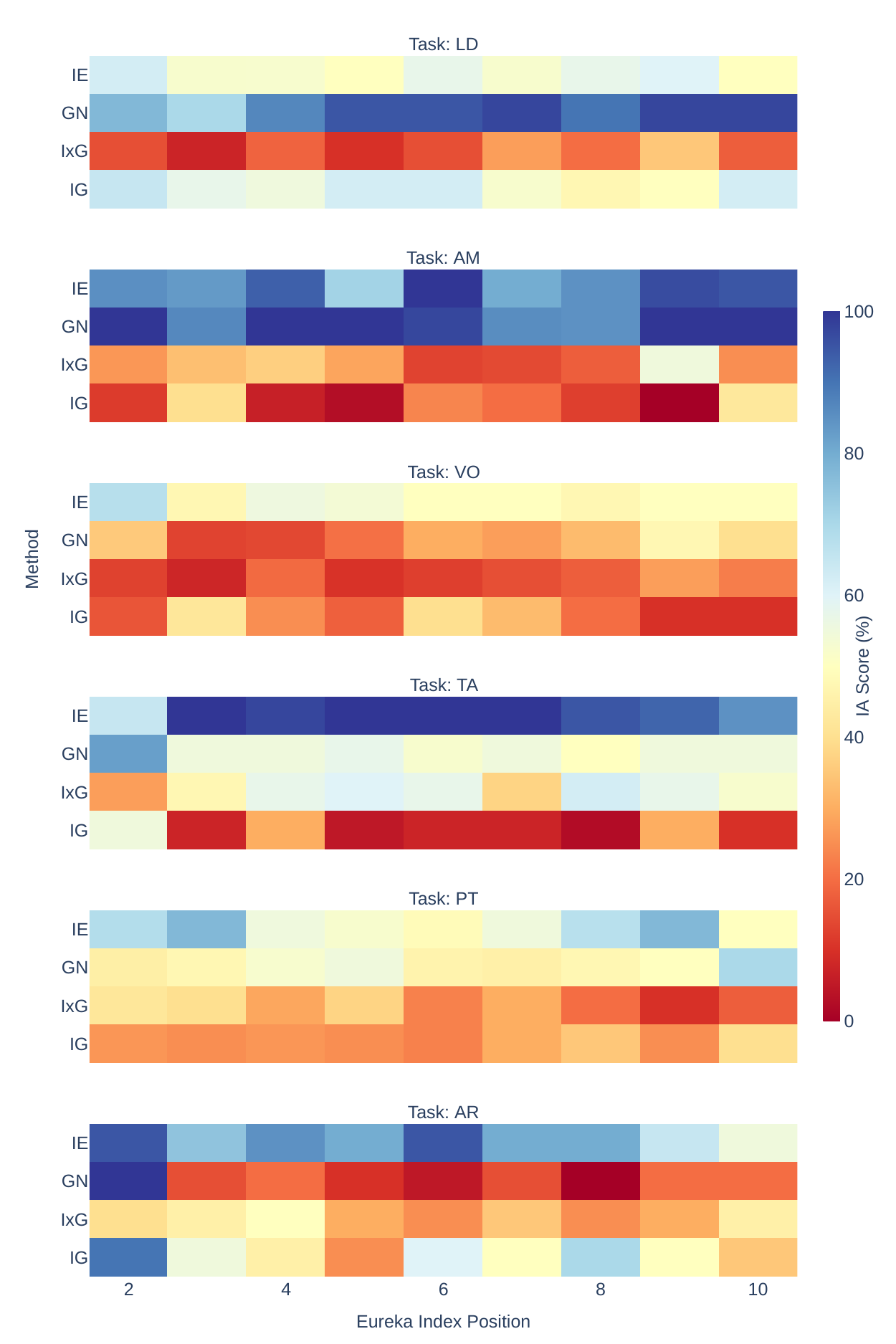}
    \caption{Distribution of the position of the example with the highest attribution scores across IA methods (Gemma-2-9B model, 10-Shot).}
    \label{fig:g9b_distribution}
\end{figure*}

\begin{figure*}[t]
    \centering
    \includegraphics[width=0.98\linewidth]{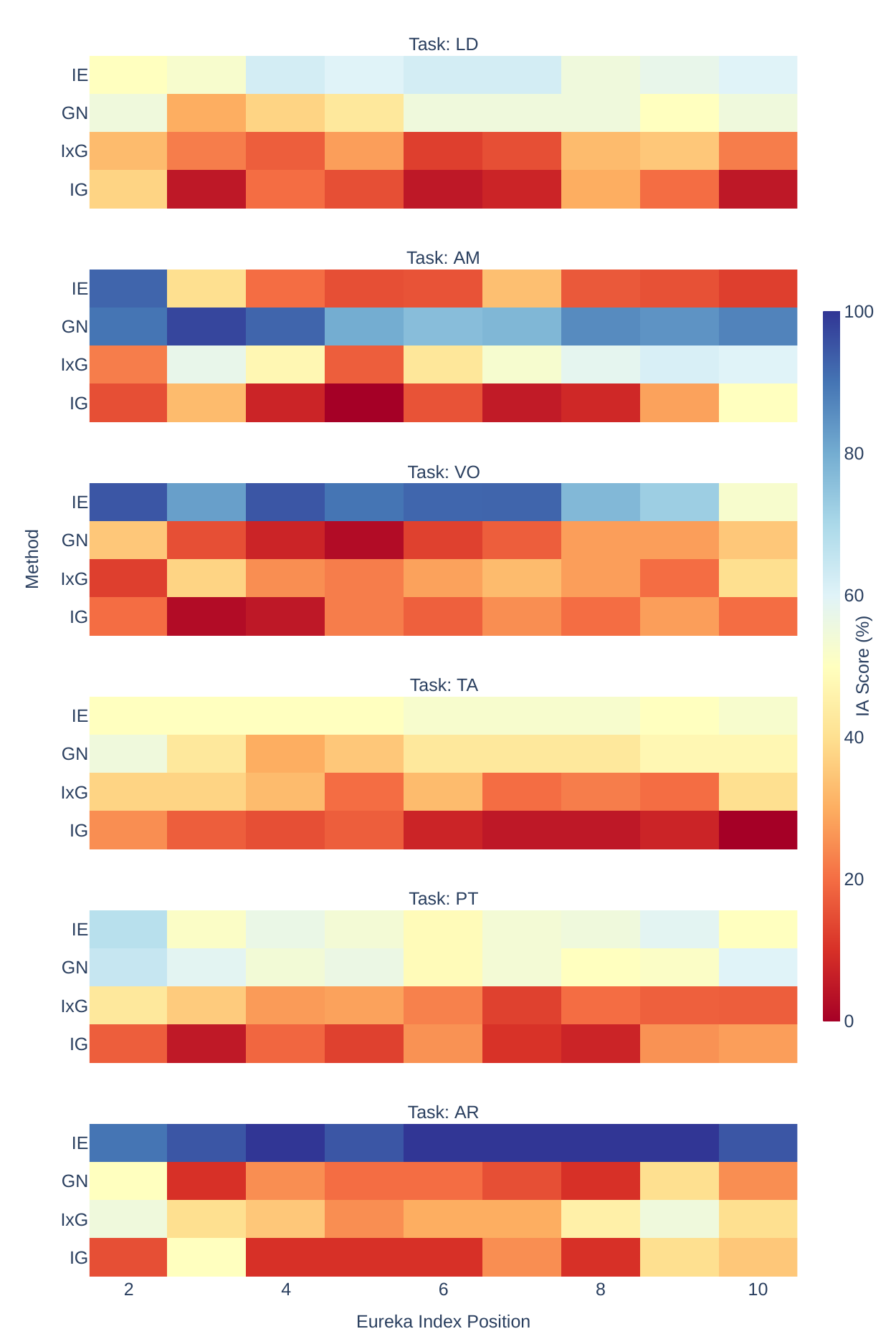}
    \caption{Distribution of the position of the example with the highest attribution scores across IA methods (Gemma-2-27B model, 10-Shot).}
    \label{fig:g27b_distribution}
\end{figure*}

\begin{figure*}[t]
    \centering
    \includegraphics[width=0.98\linewidth]{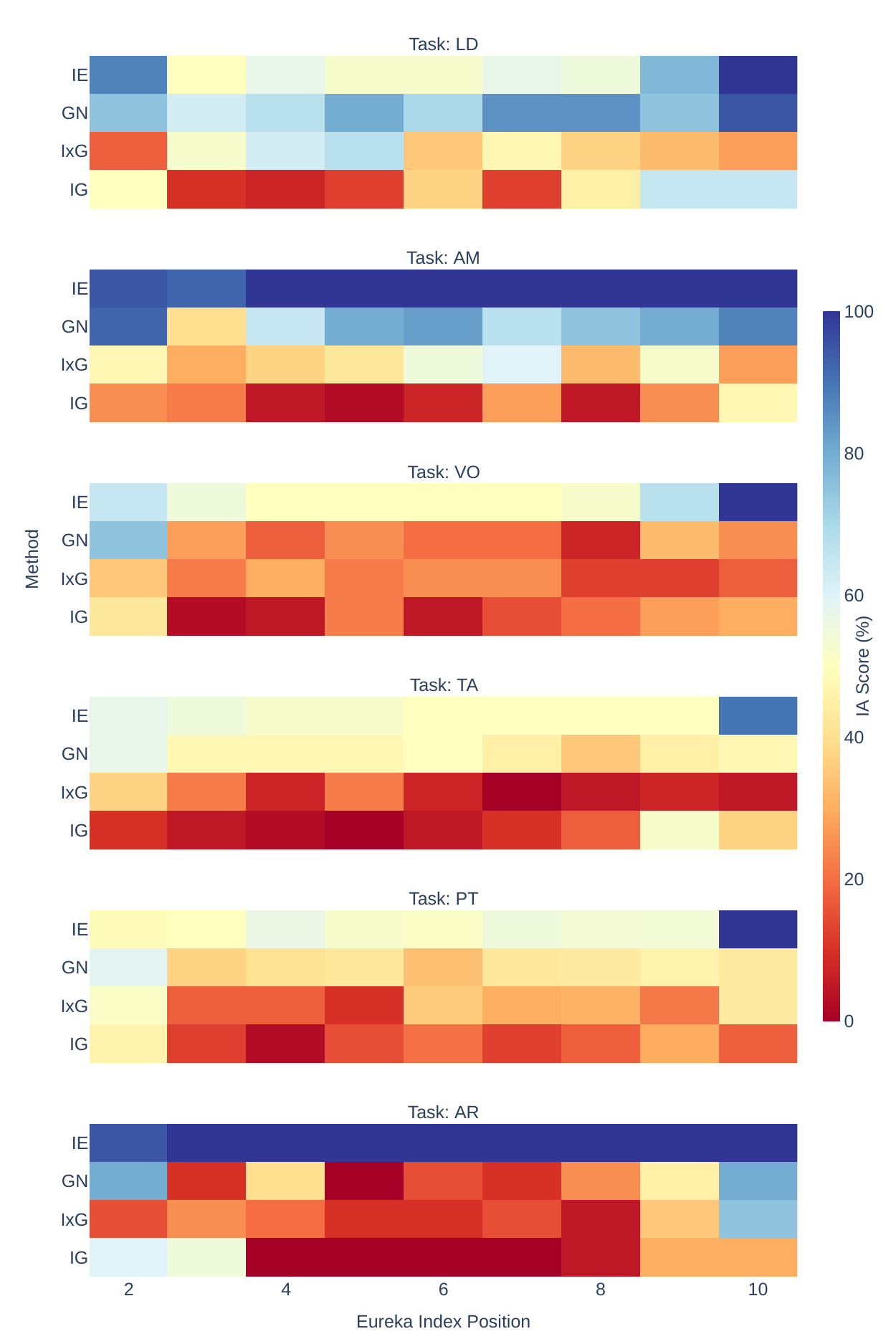}
    \caption{Distribution of the position of the example with the highest attribution scores across IA methods (Llama-2-7B model, 10-Shot).}
    \label{fig:l7b_distribution}
\end{figure*}

\begin{figure*}[t]
    \centering
    \includegraphics[width=0.98\linewidth]{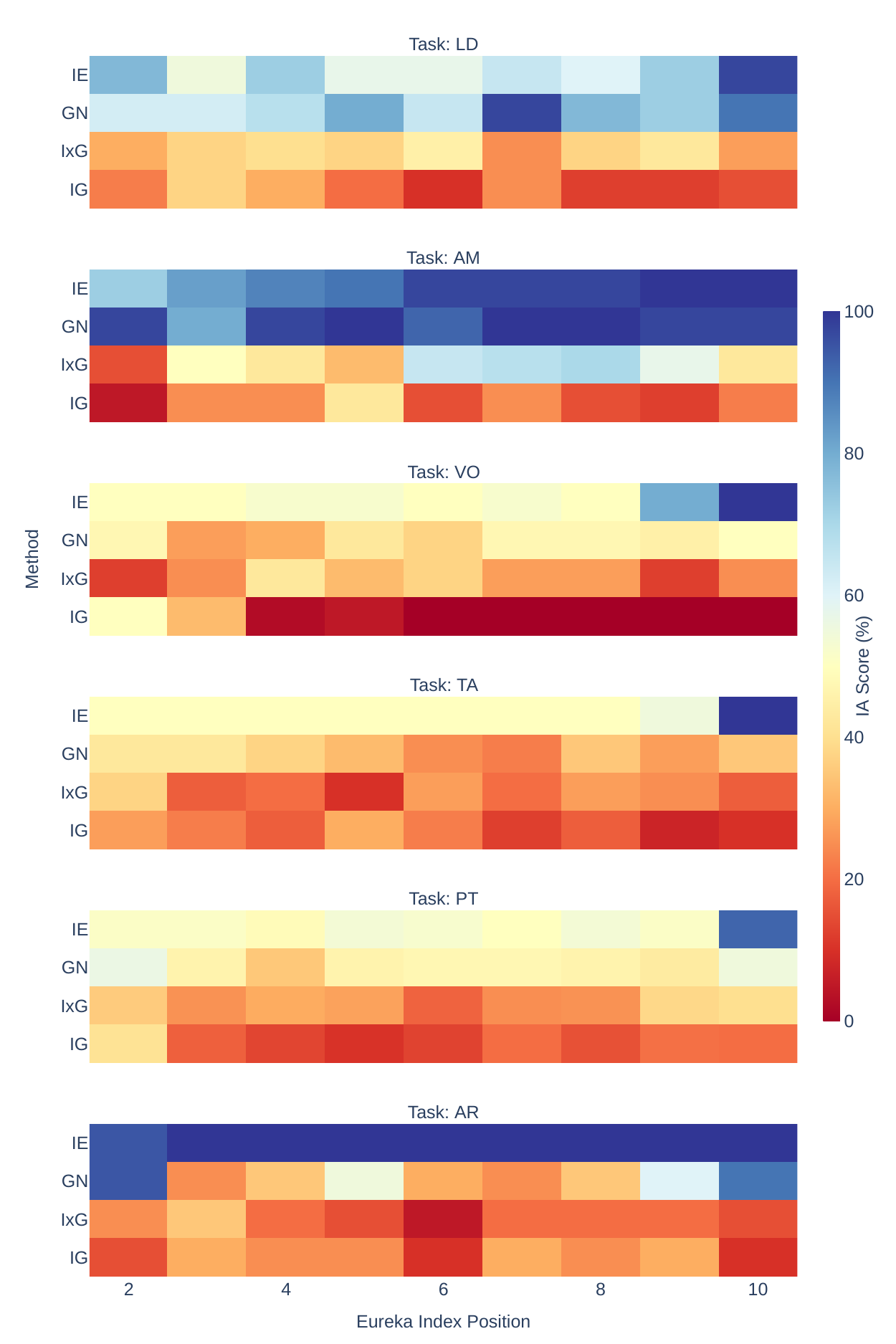}
    \caption{Distribution of the position of the example with the highest attribution scores across IA methods (Llama-2-13B model, 10-Shot).}
    \label{fig:l13b_distribution}
\end{figure*}

\begin{figure*}[t]
    \centering
    \includegraphics[width=0.98\linewidth]{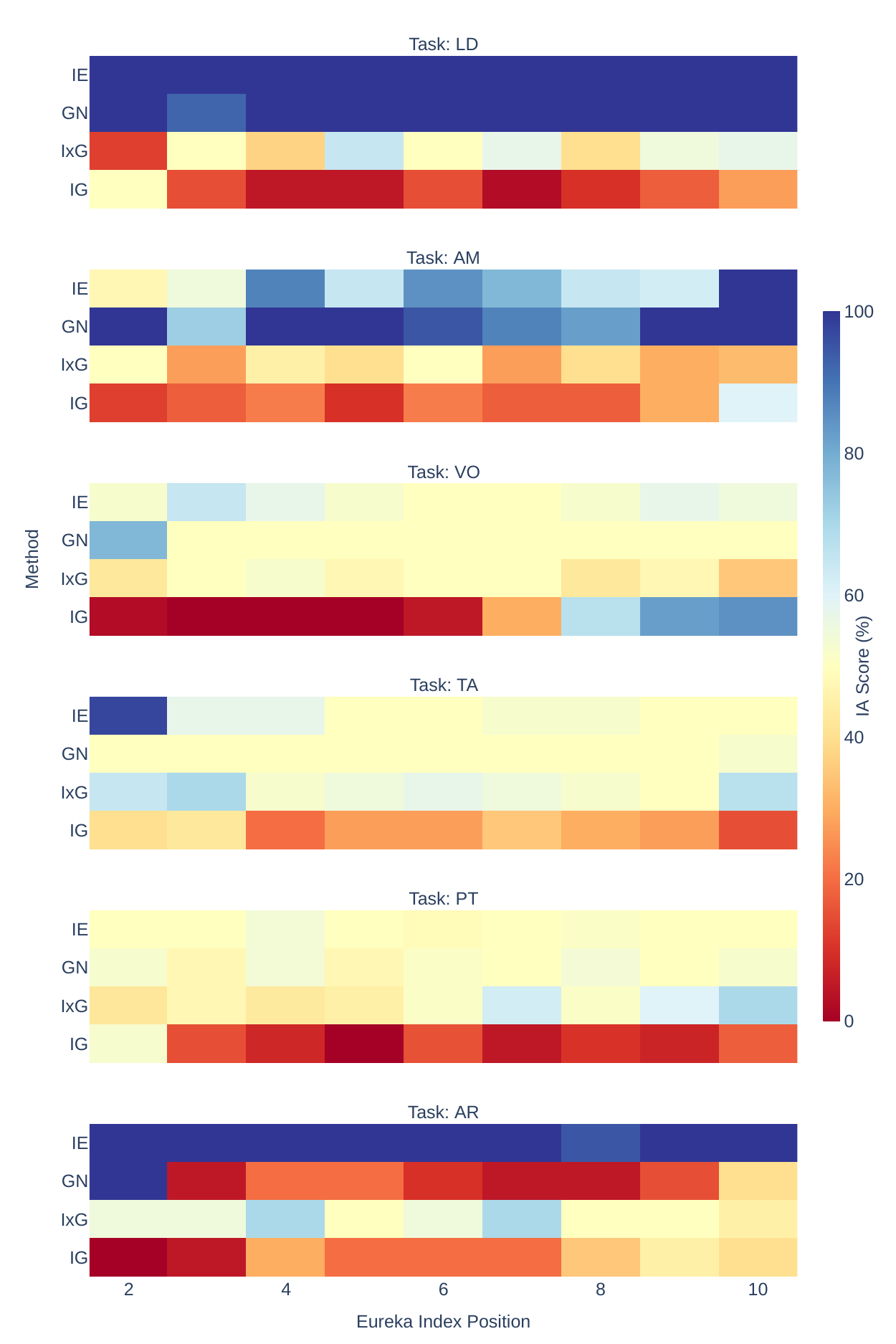}
    \caption{Distribution of the position of the example with the highest attribution scores across IA methods (Mistral-7B model, 10-Shot).}
    \label{fig:m7b_distribution}
\end{figure*}

Figures~\ref{fig:g2b_distribution_50}, \ref{fig:g2b_distribution_100}, \ref{fig:g9b_distribution}, \ref{fig:g27b_distribution}, \ref{fig:l7b_distribution}, \ref{fig:l13b_distribution} and \ref{fig:m7b_distribution} present the distribution of example positions with the highest attribution scores. All IA methods, except for I $\times$ G, possess positional bias for certain tasks, specifically favoring examples either at the beginning or end, aligning with the position bias known to LLMs~\cite{liu-etal-2024-lost}.

\section{Use of existing assets}
\label{appendix:assets}
Table~\ref{tab:used_assetss} shows the assets being used in this paper, with the type, name, link, license, and citation for each asset used in the paper.

\section{Compute statement}
Most experiments were conducted on a cluster equipped with NVIDIA A100 or H100 GPUs (80\,GB each), while some fine-tuning runs were executed on H200 GPUs with 141\,GB of memory. 
Fine-tuning time varies across models and tasks: the average run for the smallest model (Gemma-2-2B), takes roughly 2~hours per task, whereas the largest model (Gemma-2-27B), requires about 5~hours. 
IA evaluation time depends primarily on model size and the number of few-shot examples. Under the 10-shot setting, evaluation takes around 10~hours for (Gemma-2-2B) and about 24~hours for (Gemma-2-27B). 
In total, the experiments in this study consumed approximately 700 GPU hours, including exploratory runs.

\clearpage
\begin{table}[t]

\centering
\tabcolsep  4.5pt
\small
\begin{tabular}{lrr}
\toprule
\multirow{2}{*}{}
 & Zero-shot~(\%) & Few-shot~(\%) \\
\cmidrule(r){1-1} \cmidrule(lr){2-2} \cmidrule(l){3-3}
   	\textsc{Associative-Recall}  & $10.0$  & $100.0$~($10.0$)  \\
   	\textsc{Linear-or-Distinct}    & $44.8$ & $99.0$~($50.0$) \\
   	\textsc{Add-or-Multiply}  & $11.8$  & $100.0$~($50.0$) \\
   	\textsc{Verb-Object}  & $0.0$  & $100.0$~($50.0$) \\
   	\textsc{Tense-Article} & $0.0$ & $100.0$~($50.0$) \\
   	\textsc{Pos-Title} & $0.0$ & $98.0$~($50.0$) \\
\bottomrule
\end{tabular}

\caption{The zero-shot and few-shot accuracy of the fine-tuned Gemma-2-2B model across all evaluation tasks. The chance rate is indicated in parentheses.}
\label{tab:finetuning_accuracy}
\end{table}
\begin{table}[t]
\centering
\tabcolsep  4.5pt
\small
\begin{tabular}{lrr}
\toprule
\multirow{2}{*}{}
 & Zero-shot~(\%) & Few-shot~(\%) \\
\cmidrule(r){1-1} \cmidrule(lr){2-2} \cmidrule(l){3-3}
   \textsc{Associative-Recall}  & $12.0$  & $100.0$~($10.0$)  \\
   \textsc{Linear-or-Distinct}    & $50.0$ & $85.5$~($50.0$) \\
   \textsc{Add-or-Multiply}  & $12.0$  & $100.0$~($50.0$) \\
   \textsc{Verb-Object}  & $0.0$  & $94.8$~($50.0$) \\
   \textsc{Tense-Article} & $0.0$ & $100.0$~($50.0$) \\
   \textsc{Pos-Title} & $50.0$ & $98.8$~($50.0$) \\
\bottomrule
\end{tabular}

\caption{The zero-shot and few-shot accuracy of the fine-tuned Gemma-2-9B model across all evaluation tasks. The chance rate is indicated in parentheses.}
\label{tab:g2_9b_zero_acc}
\end{table}
\begin{table}[t]
\centering
\tabcolsep  4.5pt
\small
\begin{tabular}{lrr}
\toprule
\multirow{2}{*}{}
 & Zero-shot~(\%) & Few-shot~(\%) \\
\cmidrule(r){1-1} \cmidrule(lr){2-2} \cmidrule(l){3-3}
   \textsc{Associative-Recall}  & $15.0$  & $100.0$~($10.0$)  \\
   \textsc{Linear-or-Distinct}    & $47.3$ & $99.8$~($50.0$) \\
   \textsc{Add-or-Multiply}  & $48.5$  & $97.3$~($50.0$) \\
   \textsc{Verb-Object}  & $0.0$  & $99.5$~($50.0$) \\
   \textsc{Tense-Article} & $50.0$ & $100.0$~($50.0$) \\
   \textsc{Pos-Title} & $45.5$ & $97.8$~($50.0$) \\
\bottomrule
\end{tabular}

\caption{The zero-shot and few-shot accuracy of the fine-tuned Gemma-2-27B model across all evaluation tasks. The chance rate is indicated in parentheses.}
\label{tab:g2_27b_zero_acc}
\end{table}
\begin{table}[t]
\centering
\tabcolsep  4.5pt
\small
\begin{tabular}{lrr}
\toprule
\multirow{2}{*}{}
 & Zero-shot~(\%) & Few-shot~(\%) \\
\cmidrule(r){1-1} \cmidrule(lr){2-2} \cmidrule(l){3-3}
   \textsc{Associative-Recall}  & $10.0$  & $100.0$~($10.0$)  \\
   \textsc{Linear-or-Distinct}    & $44.8$ & $100.0$~($50.0$) \\
   \textsc{Add-or-Multiply}  & $11.8$  & $99.0$~($50.0$) \\
   \textsc{Verb-Object}  & $0.0$  & $100.0$~($50.0$) \\
   \textsc{Tense-Article} & $0.0$ & $100.0$~($50.0$) \\
   \textsc{Pos-Title} & $0.0$ & $98.0$~($50.0$) \\
\bottomrule
\end{tabular}

\caption{The zero-shot and few-shot accuracy of the fine-tuned Llama-2-7B model across all evaluation tasks. The chance rate is indicated in parentheses.}
\label{tab:l2_7b_zero_acc}
\end{table}
\begin{table}[t]
\centering
\tabcolsep  4.5pt
\small
\begin{tabular}{lrr}
\toprule
\multirow{2}{*}{}
 & Zero-shot~(\%) & Few-shot~(\%) \\
\cmidrule(r){1-1} \cmidrule(lr){2-2} \cmidrule(l){3-3}
   \textsc{Associative-Recall}  & $10.0$  & $100.0$~($10.0$)  \\
   \textsc{Linear-or-Distinct}    & $50.0$ & $99.8$~($50.0$) \\
   \textsc{Add-or-Multiply}  & $41.0$  & $100.0$~($50.0$) \\
   \textsc{Verb-Object}  & $0.0$  & $100.0$~($50.0$) \\
   \textsc{Tense-Article} & $0.0$ & $100.0$~($50.0$) \\
   \textsc{Pos-Title} & $41.8$ & $97.0$~($50.0$) \\
\bottomrule
\end{tabular}

\caption{The zero-shot and few-shot accuracy of the fine-tuned Llama-2-13B model across all evaluation tasks. The chance rate is indicated in parentheses.}
\label{tab:l2_13b_zero_acc}
\end{table}
\begin{table}[t]
\centering
\tabcolsep  4.5pt
\small
\begin{tabular}{lrr}
\toprule
\multirow{2}{*}{}
 & Zero-shot~(\%) & Few-shot~(\%) \\
\cmidrule(r){1-1} \cmidrule(lr){2-2} \cmidrule(l){3-3}
   \textsc{Associative-Recall}  & $0.5$  & $100.0$~($10.0$)  \\
   \textsc{Linear-or-Distinct}    & $50.0$ & $99.8$~($50.0$) \\
   \textsc{Add-or-Multiply}  & $50.0$  & $100.0$~($50.0$) \\
   \textsc{Verb-Object}  & $50.0$  & $100.0$~($50.0$) \\
   \textsc{Tense-Article} & $50.0$ & $100.5$~($50.0$) \\
   \textsc{Pos-Title} & $50.0$ & $98.3$~($50.0$) \\
\bottomrule
\end{tabular}

\caption{The zero-shot and few-shot accuracy of the fine-tuned Mistral-7B model across all evaluation tasks. The chance rate is indicated in parentheses.}
\label{tab:mistral_zero_acc}
\end{table}


\begin{table*}
\scriptsize
\centering
\begin{tabular}{ccccc}
\toprule
Asset Type & Asset Name & Link & License & Citation \\
\cmidrule(r){1-1} \cmidrule(r){2-2} \cmidrule(r){3-3} \cmidrule(r){4-4} \cmidrule(r){5-5}
Code & Contrastive Explanations & \href{https://github.com/kayoyin/interpret-lm}{\faGithub} & CC BY 4.0 & \cite{yin-neubig-2022-interpreting} \\
Code & Captum & \href{https://github.com/pytorch/captum}{\faGithub} & BSD-3-Clause License & \cite{kokhlikyan2020captum} \\
Model & Gemma-2-2B & \href{https://huggingface.co/google/gemma-2-2b}{google/gemma-2-2b} & Gemma License & \cite{riviere-etal-2024-gemma-2} \\
Model & Gemma-2-9B & \href{https://huggingface.co/google/gemma-2-9b}{google/gemma-2-9b} & Gemma License & \cite{riviere-etal-2024-gemma-2} \\
Model & Gemma-2-27B & \href{https://huggingface.co/google/gemma-2-9b}{google/gemma-2-27b} & Gemma License & \cite{riviere-etal-2024-gemma-2} \\
Model & Llama-2-7B & \href{https://huggingface.co/meta-llama/Llama-2-7b-hf}{meta-llama/Llama-2-7B} & Llama 2 Community License & \cite{touvron-etal-2023-llama2} \\
Model & Llama-2-13B & \href{https://huggingface.co/meta-llama/Llama-2-13b-hf}{meta-llama/Llama-2-13B} & Llama 2 Community License & \cite{touvron-etal-2023-llama2} \\
Model & Mistral-7B & \href{https://huggingface.co/mistralai/Mistral-7B-v0.3}{mistralai/Mistral-7B} & Apache license 2.0 & \cite{jiang2023mistral7b} \\
\bottomrule
\end{tabular}

\caption{The list of assets used in this work.}
\label{tab:used_assetss}
\end{table*}

\end{document}